\documentclass[journal]{IEEEtran}

\usepackage{amssymb}
\usepackage{url}
\usepackage{amsmath}
\usepackage{graphicx}
\usepackage[printonlyused]{acronym}
\usepackage{tabularx}
\usepackage{multirow}
\usepackage{booktabs}
\usepackage{graphics}
\usepackage{multicol}
\usepackage{subcaption}
\usepackage{float}
\usepackage{xcolor}
\usepackage{colortbl}

\DeclareMathOperator{\similarity}{sim}

\DeclareMathOperator{\t2i}{t2i}
\DeclareMathOperator{\i2t}{i2t}
\DeclareMathOperator{\rettoken}{[RET]}

\ifCLASSINFOpdf
\else
\fi

\begin{document}

\title{Large Language Models for Captioning and Retrieving Remote Sensing Images}

\author{João Daniel Silva,
        João Magalhães,
        Devis Tuia,~\IEEEmembership{Senior Member,~IEEE,}
        Bruno Martins,~\IEEEmembership{Senior Member,~IEEE}%
\thanks{João Daniel Silva and Bruno Martins are with INESC-ID, Instituto Superior Técnico, University of Lisbon, Portugal (e-mail: joao.daniel.silva@tecnico.ulisboa.pt).}%
\thanks{João Magalhães is with the Department of Computer Science, Faculty of Science and Technology, Universidade NOVA de Lisboa, Portugal.}%
\thanks{Devis Tuia is with the Ecole Polytechnique Fédérale de Lausanne, 1015 Lausanne, Switzerland.}%
\thanks{Manuscript received ---; revised ----.}}

\maketitle

\begin{abstract}
Image captioning and cross-modal retrieval %
are examples of tasks that involve the joint analysis of visual and linguistic information. In connection to remote sensing imagery, these tasks can help non-expert users in extracting relevant Earth observation information for a variety of applications. Still, despite some previous efforts, the development and application of vision and language models to the remote sensing domain have been hindered by the relatively small size of the available datasets and models used in previous studies. %
In this work, we propose RS-CapRet, a Vision and Language method for remote sensing tasks, in particular image captioning and text-image retrieval. 
We specifically propose to use a highly capable large decoder language model together with image encoders adapted to remote sensing imagery through contrastive language-image pre-training. 
To bridge together the image encoder and language decoder, we propose training simple linear layers with examples from combining different %
remote sensing image captioning datasets, keeping the other parameters frozen. RS-CapRet can then generate descriptions for remote sensing images and retrieve images from textual descriptions, achieving SOTA or competitive performance with existing methods. 
Qualitative results illustrate that RS-CapRet can effectively leverage the pre-trained large language model to describe remote sensing images, retrieve them based on different types of queries, and also show the ability to process interleaved sequences of images and text in a dialogue manner.
\end{abstract}

\begin{IEEEkeywords}
Remote Sensing Vision and Language,
Image Captioning,
Cross-modal Retrieval,
Transformers.
\end{IEEEkeywords}

\IEEEpeerreviewmaketitle

\section{Introduction}
\IEEEPARstart{T}{here} has been a growing interest in the application of Vision and Language (V\&L) models in the Remote Sensing domain~\cite{Wen2023vision,mai2023opportunities}, for tasks such as image retrieval~\cite{remoteclip,rahhal2022multilanguage,mi2022knowledge,yuan2022remote}, image captioning~\cite{cheng2022NWPUCaptions,wei2023vlca,ramos2022using,shi2017Can} and visual question answering (VQA)~\cite{lobry2020rsvqa,silva2022remote,bazi2022bi,zhang2023spatial}. Methods developed for these tasks can enable a wider population with different degrees of expertise to interact with Earth Observation data~\cite{tuia2021collective,martins2022towards}, and allow to extract richer insights from remote sensing images. %

Previous methods have adopted deep learning methods for vision and language (V\&L) tasks. Despite some recent efforts~\cite{Wen2023vision,remoteclip,hu2023rsgpt, zhan2024skyeyegpt,kuckreja2023geochat}, the relatively small size of the available datasets of image-text pairs has restrained the application and development of Vision and Language models in the remote sensing domain. These methods contrast with the current trend in general domain images, where models are getting increasingly complex and trained with large-scale datasets~\cite{wang2021SimVLM,li2022blip,alayrac2022flamingo}. In the context of tasks such as image captioning or VQA, many approaches continue to adopt an encoder-decoder architecture, leveraging CNNs as image encoders. For VQA, these methods use LSTM~\cite{hochreiter1997Long} or BERT~\cite{devlin2019BERT} to obtain a representation for the question and treat the task as a classification objective instead of generating the answer text; while for image captioning LSTMs are widely used to generate the text~\cite{chappuis2022prompt,silva2022remote,lu2017exploring,cheng2022NWPUCaptions}. However, a recent trend has included more recent Transformer based architectures and more capable vision encoders for these tasks~\cite{bazi2022bi,silva2022remote, zia2022Transforming, wei2023vlca}. Moreover, a significant amount of work done by researchers to address V\&L tasks in the remote sensing domain has focused on how to develop intricate attention mechanisms that take into account the details of this type of images~\cite{cheng2022NWPUCaptions,huang2021denoising,li2020multi,yuan2020exploring,zhao2021Highresolution}.
Still, the community recognizes the potential benefits of recent developments in Vision and Language methods applied to the remote sensing domain~\cite{mai2023opportunities}, including model architecture, pre-training methods, and their capabilities for downstream applications for the remote sensing domain.

Recent advances in state-of-the-art natural language processing (NLP) are being achieved by Large Language Models (LLM) that have been shown to have great zero-shot ability in different tasks, by performing logical reasoning and applying commonsense knowledge~\cite{petroni2019language,brown2020language,wei2022emergent,wei2022chain,kojima2022large,huang2022towards}. This observation has motivated research into how to leverage these capabilities of LLMs to integrate visual information and address V\&L tasks~\cite{guo2022images, yang2022empirical, alayrac2022flamingo, tsimpoukelli2021Multimodal}. %
However, some disadvantages of the use of these LLMs remain, mainly the high memory cost of the use of these models. Additionally, the high cost of fine-tuning these models to a specific task or application remains a big drawback. This has motivated wide research on how to efficiently adapt these models with minimum memory requirements. Proposals include different approaches for parameter-efficient training of large models, by keeping most of the parameters of the model frozen and training only a small number of newly added parameters, while still showing the ability to achieve high results and maintain the generalization abilities from the original LLM~\cite{hu2021lora,lester2021power,tsimpoukelli2021Multimodal, koh2023Grounding,eichenberg2022MAGMA, ramos2022smallcap,li2023blip}.

In this work, we propose RS-CapRet, a method that combines the high capabilities of a Large Language Model, together with an image encoder adapted to the remote sensing domain. 
Instead of fine-tuning the entire LLM and the vision encoder, we choose to freeze their parameters and only train a linear layer that is used to project the visual embeddings to the input embedding space of the decoder model, so that the visual information is in the form of an embedding vector that the LLM can process. 
We add special retrieval token $\rettoken$~\cite{koh2023Grounding}, where its embedding representation is projected into the image embeddings' space. Contrastive learning is used to train the embeddings of the $\rettoken$ token to retrieve the embeddings of the corresponding images. After the training process, this allows that an image can be retrieved given a text description, based on its similarity with the $\rettoken$ token embedding.

Despite the simplicity of the training procedure of RS-CapRet, it can generate descriptions for remote sensing images, outperforming techniques that employ complex attention mechanisms specifically tailored for remote sensing. The qualitative results demonstrate the capacity of our method to describe the contents of a remote sensing image, as well as the ability to combine image and text inputs in dialogue form and demonstrate reasoning abilities.

\section{Related Work}
Due to the increasing availability of multimodal data in the remote sensing domain, an increasing amount of work has been done tackling tasks such as image captioning or cross-modal retrieval. Recent work has introduced foundational models for the domain, including vision encoders as well as multimodal Vision and Language models.

\subsection{Image Captioning}

Previous image captioning methods in the remote sensing domain have been based on encoder-decoder architectures~\cite{qu2016Deep,lu2017exploring,zia2022Transforming, cheng2022NWPUCaptions,zhao2021Highresolution,yuan2020exploring,li2020multi,sumbul2021sdrsic,li2021truncation,huang2021denoising}, leveraging CNNs as image encoders and LSTM to generate the caption word-by-word according to weights obtained by an attention component. Most approaches have proposed mechanisms in the attention component to take into account specific characteristics of remote sensing images such as dealing with visual features at different scales~\cite{huang2021denoising,yuan2020exploring,cheng2022NWPUCaptions}. These methods are evaluated on three main datasets, namely RSICD~\cite{lu2017exploring}, Sydney-Captions~\cite{qu2016Deep}, and UCM-Captions~\cite{qu2016Deep}. MLCA-Net proposed in~\cite{cheng2022NWPUCaptions} is an example of such a model that has achieved high performance on these datasets, where a VGG backbone~\cite{Simonyan2014vgg} is used to extract features at different resolutions that are combined in multilevel and contextual attention mechanisms, which are then passed to a LSTM~\cite{hochreiter1997Long} to generate a caption. The authors have also created a new dataset for image captioning named NWPU-Captions, which has a higher quantity of data together with more diversity of descriptions and image contents. Zia et al.~\cite{zia2022Transforming} propose an encoder-decoder architecture based on a Transformer~\cite{vaswani2017Attention}, with image features obtained with a CNN developed to get features at multiple stages; they also include a topic modeling stage of the captions as input to the decoder.

Some recent work has proposed methods incorporating recent V\&L methods developed in the general domain for remote sensing. VLCA~\cite{wei2023vlca} leverage a CLIP model to obtain image features and train a cross-modal network to produce a representation to be used in a cross-attention layer of a GPT-2~\cite{radford2019Language} to generate descriptions of the image. 

\subsection{Cross-Modal Retrieval}

Remote sensing cross-modal retrieval is a task with increasing interest that can be used to evaluate representations of V\&L models. Most previous work has obtained image features with CNN and text features with LSTM or Transformer encoders, with different attention mechanisms proposed. In particular, GaLR~\cite{yuan2022remote} introduces a method that leverages both global features from a CNN and local obtained with a graph convolution network. They also apply a post-processing stage with a multivariate rerank algorithm to improve the accuracy without further training. KCR~\cite{mi2022knowledge} proposes the usage of a knowledge graph to incorporate in-domain information about the concepts mentioned in the captions, enriching the textual embeddings extracted with a SentenceBERT~\cite{reimers2019SentenceBERT}. An attention mechanism is leveraged to combine features extracted at different stages from a CNN and a triplet loss is used to optimize the model end-to-end.

CLIP~\cite{radford2021Learninga} is a V\&L model that has been trained with a contrastive loss such that images and their corresponding captions are close in the embedding space. Due to the high-quality representations of CLIP, it has motivated works also in the remote sensing domain for cross-modal retrieval. Pal et al.~\cite{pal2021Fine} finetuned CLIP with the RSICD dataset studying the impact of different augmentations both for the images and the text, and showed that the resulting model has high-quality representations, particularly for image classification. Rahhal et al.~\cite{rahhal2022multilanguage} also fine-tuned CLIP in both a single and multi-language context, obtaining high results for cross-modal retrieval. RemoteCLIP~\cite{remoteclip} has developed a pipeline to process available datasets of object detection and semantic segmentation to increase the number of image and text pairs. With this higher amount of data, they could train a CLIP model with a ViT-L backbone~\cite{dosovitskiy2020vit} and improve the results in cross-modal retrieval, compared to the ViT-B backbone used in previous approaches. TACOSS proposed by Zermatten et al.~\cite{zermatten2023tacoss} learns a fine-grained alignment between visual and textual features with a contrastive learning objective, showing to be able to perform semantic segmentation at the pixel level with this method. %

\subsection{Foundational Models for Remote Sensing}

\begin{figure*}[t!]
\centering
\includegraphics[width=\textwidth]{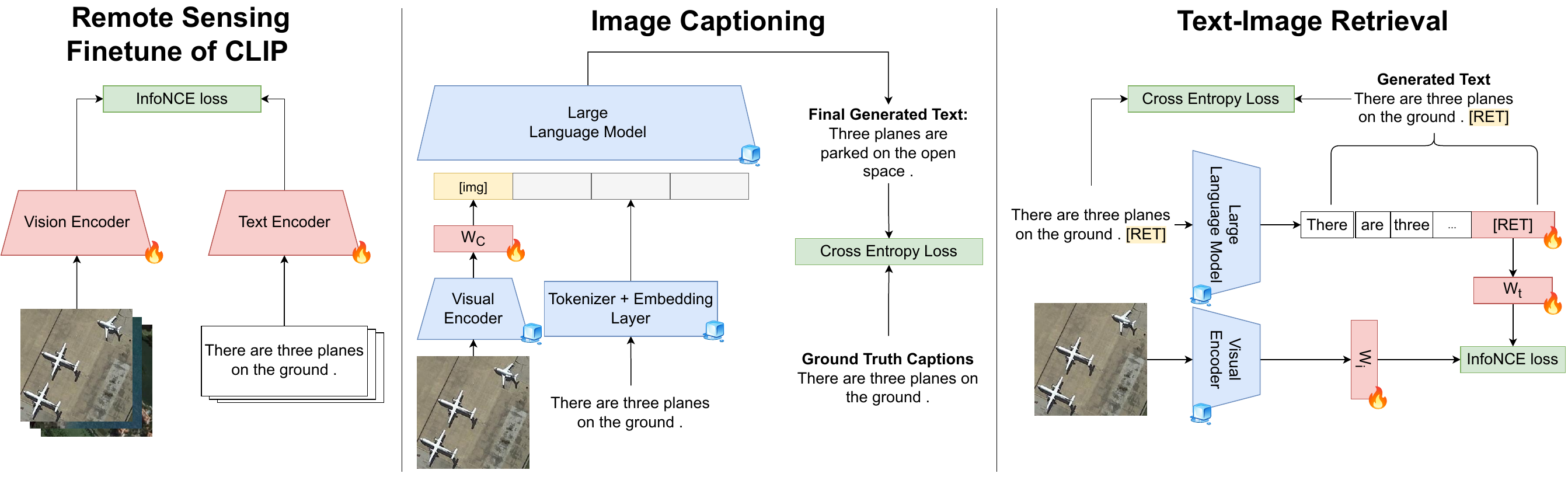}
\caption{Overview of the method used. Left: CLIP is finetuned to the remote sensing domain with image-text pairs from image captioning datasets. Middle: Image captioning task where image embeddings are obtained via a frozen image encoder and projected with a trainable linear layer to the input embedding space of the frozen large language model, which are then concatenated with the input text. Right: Trainable linear layers to apply contrastive learning between image representations and a special $\rettoken$ token to address text-image retrieval.} \label{method}
\end{figure*}

There has been a growing desire to develop a foundational model for the remote sensing community~\cite{mai2023opportunities,Wen2023vision,jakubik2023toward,martins2022towards}. A line of research has mainly focused on furthering the capabilities of vision encoders for the remote sensing domain by pre-training in a self-supervised manner with different objectives, leveraging the available high quantity of unlabeled remote sensing images. These models are then used as backbones for other methods such as object detection and semantic segmentation. Wang et al.~\cite{wang2022Advancing} pretrain a Vision Transformer with 100M parameters with a Masked Auto Encoder objective~\cite{he2022Masked} in the MillionAID dataset~\cite{Long2021DiRS}. They also propose a new rotated varied-size window attention (RVSA) with different orientation angles for computing attention. RingMO~\cite{sun2022Ringmo} is a ViT model pre-trained with a masked image modeling (MIM) objective, which the authors argue is best to address local features and tiny objects. Taking into account this type of image, the authors also argue against random masking as in usual MIM methods, since it completely loses target information in masked patches due to the tiny size of the objects in remote sensing imagery. Instead, only some pixels are randomly masked in the patches, allowing for some information to still be maintained. %
Cha et al.~\cite{cha2023billion} follow the same pre-training strategy and try to scale to models with over a billion parameters. %
Geospatial Foundational Model (GFM)~\cite{mendieta2023gfm} is an approach that combines both teacher-student distillation with masked image modeling. The authors argue that features learned from ImageNet are still helpful and can be used as auxiliary distillation objective. The resulting model surpasses both models fine-tuned from ImageNet or trained in a self-supervised manner exclusively with remote sensing imagery. The developments of vision encoders for the remote sensing domain have motivated the development of unified benchmarks for different Earth Observation tasks such as GEO-Bench~\cite{lacoste2023geobench}.

Regarding Vision and Language models for the remote sensing domain, RSGPT~\cite{hu2023rsgpt} is a recent proposal that effectively adapts the InstructBLIP~\cite{dai2023instructblip} model with a proposed dataset named RSICap of high-quality image captioning data. %
This dataset can be considered as more “dense” captioning, as it covers multiple aspects of the image such as theme, image attributes, object attributes (shape, color, quantity, size), and description of the scene. %
RSGPT excels in human evaluation on a proposed benchmark, as well as when finetuned with a few epochs in existing image captioning and visual question answering datasets~\cite{lobry2020rsvqa}. %
Kuckreja et al. proposed GeoChat~\cite{kuckreja2023geochat}, following the LLaVA1.5 architecture~\cite{liu2023improved} that can tackle different vision and language remote sensing tasks in a unified way, accepting both image-level or region-specific queries, and can also ground objects in the images by referring to their spatial coordinates. It also proposed a multimodal instruction-following dataset from current RS datasets. Another recently proposed work is SkyEyeGPT~\cite{zhan2024skyeyegpt} which also addresses multiple vision and language tasks, including visual question answering, image captioning, and vision grounding. With an architecture based on Minigpt-v2~\cite{chen2023minigpt} it achieves state-of-the-art results, in particular for image captioning and image grounding datasets.

In this work, we contribute with a Vision and Language foundational model capable of image captioning and also text-image retrieval, addressing the remote sensing domain with a simple training procedure of aligning the outputs of a visual encoder with the input space of a language model.

\section{Method}
Our work is motivated by the recent developments of leveraging Large Language Models for Vision and Language tasks. %
The proposed RS-CapRet is a model consisting of a highly capable LLM~\cite{touvron2023llama2} that can process visual inputs of the Remote Sensing domain and handle image captioning and text-image retrieval. %

Firstly, the image embeddings are obtained from a vision encoder. For the LLM to process these embeddings, they are mapped with a linear layer to the LLM input embedding space. Regarding text-image retrieval, a special $\rettoken$ token~\cite{koh2023Grounding} is added to the vocabulary of the model and linear layers are used to project this special token and the visual embeddings into a common feature space. This allows RS-CapRet to retrieve an image with the most similar embedding to the embedding of the $\rettoken$ token. Given the different characteristics of the type of images of ground-level images and the aerial images used in the remote sensing domain, we finetune a vision encoder to the remote sensing domain to obtain features that better represent this type of image. Considering that the LLM used is capable of generating high-quality text, including the type of captions that are provided in image captioning datasets, and to decrease the memory requirements of the training process as well, the parameters of the LLM are kept frozen. The parameters of the previously finetuned vision encoder are also kept frozen and the gradients are only updated for the introduced linear layers and the embeddings of the added $\rettoken$ token. %
Regarding training objectives, the model is trained on an image-captioning task, as well as a contrastive learning method between the projection of the visual encoders and $\rettoken$ token embedding.

To motivate the choice of the vision encoder, we first experimented with different models measuring their performance on cross-modal retrieval tasks as a proxy to measure the quality of the image embeddings.

\subsection{Finetuning CLIP to the Remote Sensing domain}

CLIP~\cite{radford2021Learninga} is widely used as a vision encoder for different Vision and Language models leveraging large language models, due to the high-quality embeddings that it obtains~\cite{li2023blip,liu2023visual,koh2023Grounding}. Given that CLIP was mainly trained with ground-level images and inspired by recent work such as CLIP-RSICD~\cite{pal2021Fine} and RemoteCLIP~\cite{remoteclip} that improved results for downstream tasks for the remote sensing domain, we also have finetuned CLIP with remote sensing image captioning datasets consisting of image and caption pairs. Specifically, for a dataset of $M$ image-caption pairs $\mathcal{D} = \{\mathbf{x}_i, \mathbf{y}_i\}_{i=1}^{M}$, a vision encoder obtains a representation for the image $f_\phi(\mathbf{x}_i) = \mathbf{v}_i \in \mathbf{R}^{m}$, and a text decoder obtains another for the caption $t_\theta(\mathbf{y}_i) = \mathbf{u}_i \in \mathbf{R}^{m}$. During training, the InfoNCE loss for both text-to-image $\mathcal{L}_{\t2i}$ and image-to-text $\mathcal{L}_{\i2t}$ are minimized. In a batch of $N$ examples, each pair of images and captions is considered a positive while the other elements in the batch are the negatives. Given a learnable parameter $\tau$, both losses can be formalized as:
\begin{equation} \label{infonceloss_t2i}
    \mathcal{L}_{\t2i} = - \frac{1}{N} \sum^{N}_{i=1} \left( \log \frac{\exp( \similarity(\mathbf{u}_i, \mathbf{v}_i)/\tau)}{\sum^{N}_{j=1}\exp(\similarity(\mathbf{u}_i,\mathbf{v}_i)/\tau)} \right),
\end{equation}

\begin{equation} \label{infonceloss_i2t}
    \mathcal{L}_{\i2t} = - \frac{1}{N} \sum^{N}_{i=1} \left( \log \frac{\exp( \similarity(\mathbf{v}_i,\mathbf{u}_i)/\tau)}{\sum^{N}_{j=1}\exp(\similarity(\mathbf{v}_i,\mathbf{u}_i)/\tau)} \right),
\end{equation}
where the similarity is the cosine similarity given by 
$
    \similarity(\mathbf{a},\mathbf{b}) = \exp(\mathbf{a}\cdot\mathbf{b})/ (\lVert  \mathbf{a}\cdot\mathbf{b} \rVert  \lVert \mathbf{a}\cdot\mathbf{b} \rVert).
$

\subsection{RS-CapRet Architecture}

RS-CapRet consists of the following components: i) a large language model, ii) a vision encoder finetuned to the remote sensing domain to obtain image representations, iii) a linear layer to project the obtained image embeddings to the input space of the language model, and iv) two linear layers to project, respectively, the image representations and the text representation given by a special $\rettoken$ token.  A more detailed description of each component is presented as follows:

\subsubsection{Pre-trained Language Model}
The main and larger component of our method is a decoder based on the Transformer architecture~\cite{vaswani2017Attention}, which was pre-trained on an autoregressive task to generate text. Specifically,  A sentence of text $\mathbf{y}$ is tokenized into multiple $T$ tokens $\mathbf{y}=y_1,...,y_T$, which are passed through an embedding layer to obtain a sequence of embeddings %
that is given as input to the decoder. From the output embeddings of the decoder, a linear classifier predicts the next token to be generated, and a cross-entropy between the predicted token and the ground-truth is used as signal to train the language model.  %

We experiment with %
the recently introduced %
LLamaV2-7B~\cite{touvron2023llama2}, due to its text generation performance %
as well as showing few-shot, in-context learning and reasoning abilities~\cite{radford2019Language, roberts2020How, ganguli2022Predictability}.

\subsubsection{Visual Encoder}

Given an image $\mathbf{x}$, a vision encoder based on the CLIP architecture~\cite{radford2021Learninga} is used to obtain the image representation $f_\phi(\mathbf{x}) = \mathbf{v} \in \mathbb{R}^{m}$.   %

\subsubsection{Projections between modalities}
Bridging vision and text modalities together is considered in our method in two separate directions: a) a projection of the embeddings obtained from the vision encoder to the input space of the language model, and b) a projection of both the vision embeddings and the embedding of the $\rettoken$ token to a common shared space.

Considering the first setting, the visual embeddings $\mathbf{v}$ obtained from the vision encoder cannot be directly used by the language model, since besides the dimension of the output vision vector and of the language model vector being different from each other, the features obtained by the vision encoder are not aligned with the distribution of the input space of the language model. To handle this, a linear projection $\mathbf{W}_c \in \mathbb{R}^{m\times D}$ is applied to project them to the input embedding space of the language model, resulting in a visual prefix $\mathbf{v}^T\mathbf{W}_c\in \mathbb{R}^{D}$.%

The second setting is designed for cross-modal retrieval training. The $\rettoken$ token is appended at the end of each caption and its respective embedding from the output of the language model represents the overall meaning of the sentence, $\mathbf{h}_{T}=p_\theta(y_{T}|y_1,...,y_{T-1})$. Given this output embedding, a linear layer is used to project it with $\mathbf{W}_t \in \mathbb{R}^{H \times q}$ and the visual embeddings $\mathbf{v}$ with $\mathbf{W}_i\in \mathbb{R}^{D \times q}$ to a common shared space $q < D$, so 
that contrastive learning can be applied. 

\subsection{Training procedure}

RS-CapRet is jointly trained with two tasks, image captioning and image-text retrieval. A graphical depiction of the training process can be seen in Fig.~\ref{method}.

\subsubsection{Image captioning}

The image captioning task is considered as in %
~\cite{koh2023Grounding, eichenberg2022MAGMA, tsimpoukelli2021Multimodal}: conditional generation of a caption $\mathbf{y}$ given an image $\mathbf{x}$. Specifically, an image embedding is obtained from the vision encoder and projected to the input space of the language model resulting in the visual prefix $\mathbf{v}^T\mathbf{W}_c$. This visual prefix is concatenated at the start of the text sequence token embeddings of the caption $\mathbf{y}$. The log-likelihood of the generated caption, across a batch of $N$ image-text pairs, can be formalized as:

\begin{equation}
    \mathcal{L}_c = - \frac{1}{N} \sum^{N}_{i=1} \sum^{T}_{t=1} \log p_\theta (y_t | \mathbf{v}_i^T\mathbf{W}_c, y_{i,1}, ..., y_{i,t-1}). 
\end{equation}.

To increase the robustness of the model to handle inter-leveled sequences of image and text, %
two captions are concatenated together during training for the image captioning objective. %

\begin{table*}[t!]
\centering
\caption{Description of the different remote sensing image captioning datasets used for the experiments.}
\label{table:datasets}
\begin{tabular}{lcccc}
\hline
\multicolumn{1}{c}{\textbf{Dataset}} & \textbf{\#Images} & \textbf{Image Size} & \textbf{Spatial Resolution} & \textbf{\#Total Captions}  \\ \hline %
NWPU-Captions~\cite{cheng2022NWPUCaptions} & 31,500 & $256\times256$ & $\sim$30-0.2m & 157,500 \\ %
RSICD~\cite{lu2017exploring} & 10,921 & $224\times224$ & different resolutions & 54,605 \\ %
Sydney-Captions~\cite{qu2016Deep} & 613 & $500\times500$ & 0.5m & 3,065 \\ %
UCM-Captions~\cite{qu2016Deep} & 2,100 & $256\times256$ & $\sim$0.3m & 10,500 \\ %
Cap-4 & 45,134 & $224\times224$ & different resolutions & 225,670 \\ %
RemoteCLIP & 165,745 & different sizes & different resolutions & 828,725 \\ %
\hline
\end{tabular}
\end{table*}

\begin{table*}[t!]
\centering
\caption{Retrieval performance of different CLIP variants in the RSICD dataset to motivate the choice of vision encoder. RS-CapRet leverages CLIP-ViT-L/14 finetuned with an aggregation of different remote sensing image captioning datasets - Cap-4.}
\label{table:clipperformance}
\begin{tabular}{lccccccccc}
\hline
\multicolumn{1}{c}{\multirow{2}{*}{\textbf{Method}}} & \multirow{2}{*}{\textbf{Visual Backbone}} & \multirow{2}{*}{\textbf{Finetune Data}} & \multicolumn{3}{c}{\textbf{Image-Text Retrieval}} & \multicolumn{4}{c}{\textbf{Text-Image Retrieval}} \\ \cline{4-10} 
\multicolumn{1}{c}{} &  &  & R@1 & R@5 & R@10 & R@1 & R@5 & R@10 & mR \\ \hline
CLIP~\cite{radford2021Learninga} & ViT-B & Zero-shot & 4.58 & 14.55 & 23.70 & 5.80 & 16.85 & 28.23 & 15.62 \\
CLIP~\cite{radford2021Learninga} & ViT-L & Zero-shot & 6.04 & 17.48 & 27.54 & 5.03 & 19.03 & 30.25 & 17.56 \\
CLIP-RSICD~\cite{pal2021Fine} & ViT-B & RSICD & 14.09 & 30.10 & 43.64 & 11.16 & 33.25 & 48.91 & 30.19 \\
CLIP-RSICD-L & ViT-L & RSICD & 14.27 & 32.02 & 46.39 & 12.11 & 34.97 & 50.47 & 31.70 \\
\rowcolor{lime!50}CLIP-Cap-4 & ViT-L & Cap-4 & 17.02 & 33.94 & 47.76 & 13.83 & 39.07 & 56.05 & 34.61 \\
RemoteCLIP~\cite{remoteclip} & ViT-L & RemoteCLIP dataset & 18.39 & 37.42 & 51.05 & 14.73 & 39.93 & 56.58 & 36.35 \\ \hline
\end{tabular}
\end{table*}
\subsubsection{Image-text retrieval}

Contrastive learning is also incorporated into the training procedure of RS-CapRet, which has been widely used to produce joint embeddings for images and text~\cite{radford2021Learninga,jia2021scalinga}. Specifically, contrastive learning is performed over a representation of the special retrieval $\rettoken$ token and a representation of an image. The caption $\mathbf{y}$ is given to the language model, and the output representation $\mathbf{h}=p_\theta(y_{T}|y_1,...,y_{T-1})$ is obtained together with the image embeddings from the vision encoder $\mathbf{v}=f_\phi(\mathbf{x})$. Both representations are then projected to a space of the same size with matrices $\mathbf{W}_i$ and $\mathbf{W}_t$, respectively, resulting in $\mathbf{v}'=\mathbf{v}^T\mathbf{W}_i \in \mathbb{R}^q$ and $\mathbf{u}'=\mathbf{h}^T\mathbf{W}_t \in \mathbb{R}^q$.

In a similar manner to the training process for CLIP, we optimize the InfoNCE loss considering two directions: text-to-image $\mathcal{L}_{\t2i}$ and image-to-text $\mathcal{L}_{\i2t}$, as outlined in Equations~\ref{infonceloss_t2i} and~\ref{infonceloss_i2t}, with $\mathbf{v}'_i$ and $\mathbf{u}'_i$ used as the representations for the image and text, respectively, corresponding to the \emph{i}\textsuperscript{th} image-caption pair.

The final training loss can be characterized as a weighted sum of both the image captioning and contrastive learning tasks:

\begin{equation}
    \mathcal{L} = \lambda_c\mathcal{L}_c + \lambda_r (\mathcal{L}_{\t2i} + \mathcal{L}_{\i2t}).
\end{equation}

\section{Experimental Setup}\label{experimental}

In this section, a detailed description of the datasets used is first presented, as well as the components of the architecture of RS-CapRet, namely the vision encoder used which was adapted to the remote sensing domain, the Large Language model leveraged. Finally, a description of implementation details and the metrics applied for image captioning and retrieval tasks.

\subsection{Datasets}

We leverage different remote sensing image captioning datasets for our setup. (Table~\ref{table:datasets}). RSCID dataset~\cite{lu2017exploring} is constituted by 10,921 images collected from different sources. These images were manually annotated, but many captions are duplicated to ensure 5 captions per image due to some images not reaching that count during the annotation process. UCM-Captions and Sydney-Captions were proposed by Qu et al.~\cite{qu2016Deep}; these are datasets based on scene classification datasets and were repurposed for image captioning by manual annotation.
NWPU-Captions~\cite{cheng2022NWPUCaptions} is the currently largest dataset of image and text pairs, with a total of 31,500 images. Each image has 5 manually annotated captions associated with it, for a total of 157,500 sentences. The authors intended to increase the category variety of the image and balance between classes, therefore the dataset contains 45 classes describing different land cover and land use types. The spatial resolution of the images ranges between 0.2 and 30 meters. RS-CapRet was trained by combining all the aforementioned datasets to increase the quantity of available data and the diversity of geographical scenes represented. This data is subsequently referenced in the text as Cap-4. We include the RemoteCLIP dataset~\cite{remoteclip} as a reference. This dataset is generated through a pipeline that utilizes publicly available object detection and segmentation datasets to scale the quantity of image-text pairs.

\begin{table*}[t!]
\centering
\caption{Image Captioning results on the NWPU-Captions, RSICD, UCM, and Sydney datasets.
}
\label{table:cap-all-test}
\resizebox{\textwidth}{!}{
\begin{tabular}{clcccccccccc}
\hline
\textbf{Evaluation Dataset} & \multicolumn{1}{c}{\textbf{Method}} & \textbf{Visual Encoder} & \textbf{Text Decoder} & \textbf{BLEU-1} & \textbf{BLEU-2} & \textbf{BLEU-3} & \textbf{BLEU-4} & \textbf{METEOR} & \textbf{ROUGE\textunderscore L} & \textbf{CIDEr} & \textbf{SPICE} \\ \hline
\multirow{3}{*}{NWPU} & MLCA-NET~\cite{cheng2022NWPUCaptions} & VGG16 & LSTM & 0.745 & 0.624 & 0.541 & 0.478 & 0.337 & 0.601 & 1.164 & 0.285 \\
 &  \cellcolor{lime!50} RS-CapRet & \cellcolor{lime!50}CLIP-Cap-4 & \cellcolor{lime!50}LLamaV2 & \cellcolor{lime!50}\textbf{0.871} & \cellcolor{lime!50}0.786 & \cellcolor{lime!50}0.713 & \cellcolor{lime!50}0.650 & \cellcolor{lime!50}\textbf{0.439} & \cellcolor{lime!50}0.775 & \cellcolor{lime!50}1.919 & \cellcolor{lime!50}\textbf{0.320} \\
 & \cellcolor{lime!50} RS-CapRet$_{finetuned}$ & \cellcolor{lime!50}CLIP-Cap-4 & \cellcolor{lime!50}LLamaV2 & \cellcolor{lime!50}\textbf{0.871} & \cellcolor{lime!50}\textbf{0.787} & \cellcolor{lime!50}\textbf{0.717} & \cellcolor{lime!50}\textbf{0.656} & \cellcolor{lime!50}0.436 & \cellcolor{lime!50}\textbf{0.776} & \cellcolor{lime!50}\textbf{1.929} & \cellcolor{lime!50}0.311 \\ \hline
\multirow{5}{*}{RSICD} & MLCA-NET~\cite{cheng2022NWPUCaptions} & VGG16 & LSTM & 0.757 & 0.634 & 0.539 & 0.461 & 0.351 & 0.646 & 2.356 & 0.444 \\
 & RSGPT~\cite{hu2023rsgpt} & EVA-G & Vicuna & 0.703 & 0.542 & 0.440 & 0.368 & 0.301 & 0.533 & 1.029 & NA \\
 & SkyEyeGPT~\cite{zhan2024skyeyegpt} & EVA-G & LLamaV2-Chat & \textbf{0.867} & \textbf{0.767} & \textbf{0.673} & \textbf{0.600} & 0.354 & 0.626 & 0.837 & NA \\
 &  \cellcolor{lime!50}RS-CapRet & \cellcolor{lime!50}CLIP-Cap-4 & \cellcolor{lime!50}LLamaV2 & \cellcolor{lime!50}0.741 & \cellcolor{lime!50}0.622 & \cellcolor{lime!50}0.529 & \cellcolor{lime!50}0.455 & \cellcolor{lime!50}\textbf{0.376} & \cellcolor{lime!50}\textbf{0.649} & \cellcolor{lime!50}\textbf{2.605} & \cellcolor{lime!50}\textbf{0.484} \\
 &  \cellcolor{lime!50}RS-CapRet$_{finetuned}$ & \cellcolor{lime!50}CLIP-Cap-4 & \cellcolor{lime!50}LLamaV2 & \cellcolor{lime!50}0.720 & \cellcolor{lime!50}0.599 & \cellcolor{lime!50}0.506 & \cellcolor{lime!50}0.433 & \cellcolor{lime!50}0.370 & \cellcolor{lime!50}0.633 & \cellcolor{lime!50}2.502 & \cellcolor{lime!50}0.474 \\ \hline
\multirow{5}{*}{UCM} & MLCA-NET~\cite{cheng2022NWPUCaptions} & VGG16 & LSTM & 0.826 & 0.770 & 0.717 & 0.668 & 0.435 & 0.772 & 3.240 & 0.473 \\
 & RSGPT~\cite{hu2023rsgpt} & EVA-G~\cite{fang2023eva} & Vicuna~\cite{vicuna2023} & 0.861 & 0.791 & 0.723 & 0.657 & 0.422 & 0.783 & 3.332 & NA \\
 & SkyEyeGPT~\cite{zhan2024skyeyegpt} & EVA-G~\cite{fang2023eva} & LLamaV2-Chat & \textbf{0.907} & \textbf{0.857} & \textbf{0.816} & \textbf{0.784} & 0.462 & 0.795 & 2.368 & NA \\
 &  \cellcolor{lime!50}RS-CapRet & \cellcolor{lime!50}CLIP-Cap-4 & \cellcolor{lime!50}LLamaV2 & \cellcolor{lime!50}0.833 & \cellcolor{lime!50}0.760 & \cellcolor{lime!50}0.699 & \cellcolor{lime!50}0.645 & \cellcolor{lime!50}0.447 & \cellcolor{lime!50}0.786 & \cellcolor{lime!50}3.429 & \cellcolor{lime!50}\textbf{0.525} \\
 &  \cellcolor{lime!50}RS-CapRet$_{finetuned}$ & \cellcolor{lime!50}CLIP-Cap-4 & \cellcolor{lime!50}LLamaV2 & \cellcolor{lime!50}0.843 & \cellcolor{lime!50}0.779 & \cellcolor{lime!50}0.722 & \cellcolor{lime!50}0.670 & \cellcolor{lime!50}\textbf{0.472} & \cellcolor{lime!50}\textbf{0.817} & \cellcolor{lime!50}\textbf{3.548} & \cellcolor{lime!50}\textbf{0.525} \\ \hline
\multirow{5}{*}{Sydney} & MLCA-NET~\cite{cheng2022NWPUCaptions} & VGG16 & LSTM & 0.831 & 0.742 & 0.659 & 0.580 & 0.390 & 0.711 & 2.324 & 0.409 \\
 & RSGPT~\cite{hu2023rsgpt} & EVA-G & Vicuna & 0.823 & 0.753 & 0.686 & 0.622 & 0.414 & 0.748 & \textbf{2.731} & NA \\
 & SkyEyeGPT~\cite{zhan2024skyeyegpt} & EVA-G & LLamaV2-Chat & \textbf{0.919} & \textbf{0.856} & \textbf{0.809} & \textbf{0.774} & \textbf{0.466} & \textbf{0.777} & 1.811 & NA \\
 &  \cellcolor{lime!50}RS-CapRet & \cellcolor{lime!50}CLIP-Cap-4 & \cellcolor{lime!50}LLamaV2 & \cellcolor{lime!50}0.782 & \cellcolor{lime!50}0.688 & \cellcolor{lime!50}0.611 & \cellcolor{lime!50}0.545 & \cellcolor{lime!50}0.383 & \cellcolor{lime!50}0.704 & \cellcolor{lime!50}2.390 & \cellcolor{lime!50}0.423 \\
 &  \cellcolor{lime!50}RS-CapRet$_{finetuned}$ & \cellcolor{lime!50}CLIP-Cap-4 & \cellcolor{lime!50}LLamaV2 & \cellcolor{lime!50}0.787 & \cellcolor{lime!50}0.700 & \cellcolor{lime!50}0.628 & \cellcolor{lime!50}0.564 & \cellcolor{lime!50}0.388 & \cellcolor{lime!50}0.707 & \cellcolor{lime!50}2.392 & \cellcolor{lime!50}\textbf{0.434} \\ \hline
\end{tabular}
}
\end{table*}

\subsection{Vision Encoder}

The vision encoder is based on a CLIP vision model due to its high performance in different tasks and following previous work that has successfully combined it with other LLMs for Vision and Language tasks~\cite{koh2023Grounding,liu2023visual}, as well as its effectiveness for cross-modal retrieval tasks in the remote sensing domain~\cite{radford2021Learninga,remoteclip,pal2021Fine}.
Specifically, we use a CLIP of size Large finetuned with an aggregation of image captioning datasets (Cap-4), and this model is refered as CLIP-Cap-4. We further motivate our specific choice of CLIP pre-trained weights with experimental results in cross-modal retrieval, detailed in the next section.

\subsection{Language Model}

As for the decoder model, we used the recently proposed LLamaV2-7B language model~\cite{touvron2023LLaMA}, which has incorporated recent advances in NLP~\cite{hoffmann2022training}, %
and compared with the original LLaMa architecture~\cite{touvron2023llama2} is trained with more data, and has a larger context window.

\subsection{Implementation Details}

Our experiments were implemented in PyTorch~\cite{paszke2019pytorch} and trained with mixed-precision in bfloat16, to have a lower cost of training. %
The batch size was set to 64, the learning rate was set to 0.0003, with a warmup of 100 steps, and the Adam optimizer was used. Both the loss weights of the image captioning and contrastive learning tasks were equal to one, $\lambda_c=\lambda_r=1$. For simplicity, we only considered one token embedding for visual information. The dimensionality of the contrastive learning space was set to $q=256$. 
The gradient updates were only made on the parameters of the linear layers introduced and the $\rettoken$ embedding token. As for the input images size, a resize was applied to $224\times224$.

\subsection{Metrics}

We report the results of RS-CapRet for the image captioning task on the test splits of NWPU-Captions, RSICD, Sydney-Captions, and UCM datasets. Evaluation is based on common image captioning metrics following previous work such a BLEU, METEOR, ROUGE\_L, CIDEr, and SPICE.

Regarding the retrieval task, the retrieval recall of top-1 (R@1), top-5 (R@5), and top-10 (R@10), %
following previous work~\cite{mi2022knowledge, remoteclip}.
R@K means the ratio of queries that successfully retrieve the ground truth as one of the first K results. %
We do the text-image retrieval evaluation, factoring in calculating the mean recall (mR\textunderscore T2I) for the other methods, based on the values reported for an accurate comparison.

\section{Experimental Results}

\begin{table*}[t!]
\centering
\caption{Results of retrieval experiments in the RSICD and UCM datasets. RS-CapRet in text-image retrieval direction together with the other methods. Methods with~\dag~were evaluated in our setup, otherwise they are collected from the respective report. Methods in light-blue are \textcolor{cyan}{\textbf{CLIP-based}} methods, while \textcolor{lime}{\textbf{RS-CapRet}} is in light-green. %
}
\label{table:retrieval-ucm}
\begin{tabular}{clcccccc}
\hline
\multicolumn{1}{l}{\multirow{2}{*}{\textbf{Dataset}}} & \multicolumn{1}{c}{\multirow{2}{*}{\textbf{Method}}} & \multirow{2}{*}{\textbf{Visual Backbone}} & \multirow{2}{*}{\textbf{Finetune Data}} & \multicolumn{4}{c}{\textbf{Text-Image Retrieval}}  \\ \cline{5-8} 
\multicolumn{1}{l}{} & \multicolumn{1}{c}{} &  &  & R@1 & R@5 & R@10 & mR\_T2I \\ \hline
\multirow{10}{*}{RSICD} & GaLR~\cite{yuan2022remote} & ResNet18 & RSICD & 4.69 & 19.48 & 32.13 & 18.77 \\
 & KCR~\cite{mi2022knowledge} & ResNet101 & RSICD & 5.40 & 22.44 & 37.36 & 21.73 \\
 & \cellcolor{cyan!50}CLIP~\cite{radford2021Learninga}\dag & \cellcolor{cyan!50}ViT-B & \cellcolor{cyan!50}Zero-shot & \cellcolor{cyan!50}5.80 & \cellcolor{cyan!50}16.85 & \cellcolor{cyan!50}28.23 & \cellcolor{cyan!50}16.96 \\
 & \cellcolor{cyan!50}CLIP~\cite{radford2021Learninga}\dag & \cellcolor{cyan!50}ViT-L & \cellcolor{cyan!50}Zero-shot & \cellcolor{cyan!50}5.03 & \cellcolor{cyan!50}19.03 & \cellcolor{cyan!50}30.25 & \cellcolor{cyan!50}18.10 \\
 & \cellcolor{cyan!50}Rahhal et al.~\cite{rahhal2022multilanguage} & \cellcolor{cyan!50}ViT-B & \cellcolor{cyan!50}RSICD & \cellcolor{cyan!50}9.14 & \cellcolor{cyan!50}28.96 & \cellcolor{cyan!50}44.59 & \cellcolor{cyan!50}27.56 \\
 & \cellcolor{cyan!50}CLIP-RSICD~\cite{pal2021Fine}\dag & \cellcolor{cyan!50}ViT-B  & \cellcolor{cyan!50}RSICD & \cellcolor{cyan!50}11.16 & \cellcolor{cyan!50}33.25 & \cellcolor{cyan!50}48.91 & \cellcolor{cyan!50}31.11 \\
 & \cellcolor{cyan!50}CLIP-Cap-4\dag & \cellcolor{cyan!50}ViT-L & \cellcolor{cyan!50}Cap-4 & \cellcolor{cyan!50}13.83 & \cellcolor{cyan!50}39.07 & \cellcolor{cyan!50}56.05 & \cellcolor{cyan!50}36.32 \\
 & \cellcolor{cyan!50}RemoteCLIP~\cite{remoteclip} & \cellcolor{cyan!50}ViT-L & \cellcolor{cyan!50}RemoteCLIP dataset & \cellcolor{cyan!50}\textbf{14.73} & \cellcolor{cyan!50}\textbf{39.93} & \cellcolor{cyan!50}\textbf{56.58} & \cellcolor{cyan!50}\textbf{37.08} \\
 &  \cellcolor{lime!50}RS-CapRet\dag & \cellcolor{lime!50}ViT-L & \cellcolor{lime!50}Cap-4 & \cellcolor{lime!50}9.83 & \cellcolor{lime!50}30.17 & \cellcolor{lime!50}47.43 & \cellcolor{lime!50}29.14 \\
 &  \cellcolor{lime!50}RS-CapRet$_{finetuned}\dag$ & \cellcolor{lime!50}ViT-L & \cellcolor{lime!50}Cap-4 + RSICD & \cellcolor{lime!50}10.25 & \cellcolor{lime!50}31.62 & \cellcolor{lime!50}48.53 & \cellcolor{lime!50}30.13 \\ \hline
\multirow{10}{*}{UCM} & KCR~\cite{mi2022knowledge} & ResNet101 & RSICD & 17.43 & 57.52 & 80.38 & 51.78 \\
 & \cellcolor{cyan!50}CLIP~\cite{radford2021Learninga}\dag & \cellcolor{cyan!50}ViT-B & \cellcolor{cyan!50}Zero-shot & \cellcolor{cyan!50}8.67 & \cellcolor{cyan!50}36.48 & \cellcolor{cyan!50}60.57 & \cellcolor{cyan!50}35.24 \\
 & \cellcolor{cyan!50}CLIP~\cite{radford2021Learninga}\dag & \cellcolor{cyan!50}ViT-L  & \cellcolor{cyan!50}Zero-shot & \cellcolor{cyan!50}10.76 & \cellcolor{cyan!50}46.00 & \cellcolor{cyan!50}73.33 & \cellcolor{cyan!50}43.37 \\
 & \cellcolor{cyan!50}CLIP-RSICD~\cite{pal2021Fine}\dag & \cellcolor{cyan!50}ViT-B & \cellcolor{cyan!50}RSICD & \cellcolor{cyan!50}13.81 & \cellcolor{cyan!50}57.05 & \cellcolor{cyan!50}91.24 & \cellcolor{cyan!50}54.03 \\
 & \cellcolor{cyan!50}CLIP-Cap-4\dag & \cellcolor{cyan!50}ViT-L & \cellcolor{cyan!50}Cap-4 & \cellcolor{cyan!50}16.29 & \cellcolor{cyan!50}60.57 & \cellcolor{cyan!50}94.76 & \cellcolor{cyan!50}57.21 \\
 & \cellcolor{cyan!50}RemoteCLIP~\cite{remoteclip} & \cellcolor{cyan!50}ViT-L & \cellcolor{cyan!50}RemoteCLIP dataset & \cellcolor{cyan!50}\textbf{17.71} & \cellcolor{cyan!50}62.19 & \cellcolor{cyan!50}\textbf{93.90} & \cellcolor{cyan!50}57.93 \\
 & \cellcolor{cyan!50}Rahhal et al.~\cite{rahhal2022multilanguage} & \cellcolor{cyan!50}ViT-B & \cellcolor{cyan!50}UCM & \cellcolor{cyan!50}19.33 & \cellcolor{cyan!50}\textbf{64.00} & \cellcolor{cyan!50}91.42 & \cellcolor{cyan!50}\textbf{58.25} \\
 &  \cellcolor{lime!50}RS-CapRet\dag & \cellcolor{lime!50}ViT-L & \cellcolor{lime!50}Cap-4 & \cellcolor{lime!50}15.52 & \cellcolor{lime!50}57.24 & \cellcolor{lime!50}88.76 & \cellcolor{lime!50}53.84 \\
 &  \cellcolor{lime!50}RS-CapRet$_{finetuned}\dag$\dag & \cellcolor{lime!50}ViT-L & \cellcolor{lime!50}Cap-4 + UCM & \cellcolor{lime!50}16.10 & \cellcolor{lime!50}56.29 & \cellcolor{lime!50}90.76 & \cellcolor{lime!50}54.38 \\ \hline
\end{tabular}
\end{table*}

\subsection{Choice of CLIP Vision Encoder} \label{subsection:clipchoice}
We first measure the retrieval performance of different CLIP vision encoder variants as a proxy to motivate our choice of vision encoder, and the obtained results are compiled in Table~\ref{table:clipperformance}, in the RSICD dataset. %
The larger variant of CLIP ViT-L/14 obtains higher results when compared to the ViT-B/32 variant. When measuring the results of an open-source version of a CLIP ViT-B/32 fine-tuned on RSICD, named CLIP-RSICD~\cite{pal2021Fine}, the results are highly increased, even surpassing those of the larger variant. We note that the authors of this model~\cite{pal2021Fine} used extensive augmentation strategies (both for images and caption text) as well as a high batch size, which highly benefits these contrastive learning-based approaches~\cite{radford2021Learninga,chen2020Simple}. Motivated by both these results, we have also fine-tuned a CLIP ViT-L/14 with RSICD data. %
When using the Cap-4 training data instead, the model, which we named CLIP-Cap-4, further improved the results over the aforementioned CLIP variants on both RSICD and UCM datasets. From these results, we fix the vision encoder of RS-CapRet to CLIP ViT-L/14 finetuned with Cap-4 data. For completeness, we also include the results of retrieval of the newly introduced RemoteCLIP~\cite{remoteclip} which leveraged an automatic procedure to scale the training data over +800k image-text pairs. Despite its great performance we still use the CLIP-Cap-4 model because we had full control of its training procedure, which was done over only image-captioning datasets.

\subsection{Image Captioning Results}
The results for the image captioning task are compiled in Table~\ref{table:cap-all-test}. We compare our results with MLCA-NET~\cite{cheng2022NWPUCaptions} which had obtained SOTA results in image captioning at the time it was proposed and remains a strong baseline for these datasets. %
We also include the recently introduced RSGPT~\cite{hu2023rsgpt}, except for NWPU-Captions on which this model was not yet evaluated on. %
RSGPT is a finetuned version of InstructBLIP~\cite{dai2023instructblip} for the remote sensing domain, which has more parameters compared to the other baseline method MLCA-NET. 

For the NWPU-Captions dataset, RS-CapRet greatly improves over the previous SOTA model MLCA-NET~\cite{cheng2022NWPUCaptions} (e.g. +0.126 for BLEU-1, +0.755 for CIDEr). Considering the results in the RSICD dataset, RS-CapRet surpasses RSGPT in all metrics (in particular CIDEr by a high amount of +1.576). When compared to MLCA-Net and SkyEyeGPT, RS-CapRet has higher scores on all metrics but BLEU. For the UCM-Captions dataset, RS-CapRet has lower results for the BLEU metrics but achieves higher METEOR, CIDEr and SPICE, and similar ROUGE\_L when compared to RSGPT. Compared to SkyEyeGPT, RS-CapRet surpasses only in CIDEr. Regarding the results in the Sydney-Captions dataset, our results are lower compared to the baselines, and RS-CapRet only has better results in CIDEr (and SPICE compared to MLCA-NET). 
We note that this latter dataset is more specific compared to the latter ones, both in image contents (images taken over Sydney) and textual descriptions, divided into only 7 different types of classes. RS-CapRet was trained with the Cap-4 dataset, where the number of total captions from Sydney-Captions dataset corresponds to $\sim1\%$ of the total captions (see Table~\ref{table:datasets}). Thus, our model does not have much data to align its generation outputs with descriptions that correspond more to the way they are expected for Sydney-Captions; however, it still achieves a good performance. %
Overall, from these results, it can be concluded that RS-CapRet is a single model that can achieve high performance on heterogeneous remote sensing image captioning datasets.

\begin{table*}[t!]
\centering
\caption{Comparison of results in image captioning and text-image retrieval by changing the vision encoder of RS-RetCap from CLIP to one based on MAE~\cite{wang2022Advancing}, and considering encoders of different sizes. LLamaV2 was chosen as the language model as in the original architecture and the whole model was trained on Cap-4 data, following the training procedure of RS-CapRet.}
\label{table:ablation-IC}
\begin{tabular}{clccccccccc}
\hline
\multirow{2}{*}{\textbf{Dataset}} & \multicolumn{1}{c}{\multirow{2}{*}{\textbf{Visual Encoder}}} & \multirow{2}{*}{\textbf{Visual Backbone}} & \multicolumn{4}{c}{\textbf{Image Captioning}} & \multicolumn{4}{c}{\textbf{Text-Image Retrieval}} \\ \cline{4-11} 
 & \multicolumn{1}{c}{} &  & BLEU-1 & BLEU-4 & CIDEr & SPICE & R@1 & R@5 & R@10 & mR \\ \hline
\multirow{3}{*}{NWPU-Captions} & RS-ViT-B~\cite{wang2022Advancing} & ViT-B & 0.810 & 0.547 & 1.542 & 0.269 &  &  &  &  \\
 & CLIP-RSICD~\cite{pal2021Fine} & ViT-B & 0.826 & 0.565 & 1.645 & 0.276 &  &  &  &  \\
 & \cellcolor{lime!50}CLIP-Cap-4 & \cellcolor{lime!50}ViT-L & \cellcolor{lime!50}0.871 & \cellcolor{lime!50}0.650 & \cellcolor{lime!50}1.919 & \cellcolor{lime!50}0.320 &  &  &  &  \\ \hline
\multirow{3}{*}{RSICD} & RS-ViT-B~\cite{wang2022Advancing} & ViT-B & 0.706 & 0.410 & 2.329 & 0.449 & 5.00 & 17.97 & 30.01 & 17.66 \\
 & CLIP-RSICD~\cite{pal2021Fine} & ViT-B & 0.728 & 0.439 & 2.524 & 0.466 & 7.47 & 25.36 & 40.73 & 24.52 \\
 & \cellcolor{lime!50}CLIP-Cap-4 & \cellcolor{lime!50}ViT-L & \cellcolor{lime!50}0.741 & \cellcolor{lime!50}0.455 & \cellcolor{lime!50}2.605 & \cellcolor{lime!50}0.484 & \cellcolor{lime!50}9.83 & \cellcolor{lime!50}30.17 & \cellcolor{lime!50}47.43 & \cellcolor{lime!50}29.14 \\ \hline
\multirow{3}{*}{UCM} & RS-ViT-B~\cite{wang2022Advancing} & ViT-B & 0.699 & 0.467 & 2.347 & 0.374 & 9.71 & 41.43 & 69.81 & 40.32 \\
 & CLIP-RSICD~\cite{pal2021Fine} & ViT-B & 0.810 & 0.606 & 2.901 & 0.439 & 11.14 & 48.48 & 83.33 & 47.65 \\
 & \cellcolor{lime!50}CLIP-Cap-4 & \cellcolor{lime!50}ViT-L & \cellcolor{lime!50}0.833 & \cellcolor{lime!50}0.645 & \cellcolor{lime!50}3.429 & \cellcolor{lime!50}0.525 & \cellcolor{lime!50}15.52 & \cellcolor{lime!50}57.24 & \cellcolor{lime!50}88.76 & \cellcolor{lime!50}53.84 \\ \hline
\multirow{3}{*}{Sydney-Captions} & RS-ViT-B~\cite{wang2022Advancing} & ViT-B & 0.757 & 0.520 & 2.114 & 0.408 &  &  &  &  \\
 & CLIP-RSICD~\cite{pal2021Fine} & ViT-B & 0.772 & 0.538 & 2.177 & 0.406 &  &  &  &  \\
 & \cellcolor{lime!50}CLIP-Cap-4 & \cellcolor{lime!50}ViT-L & \cellcolor{lime!50}0.782 & \cellcolor{lime!50}0.545 & \cellcolor{lime!50}2.390 & \cellcolor{lime!50}0.423 &  &  &  &  \\ \hline
\end{tabular}
\end{table*}

\subsubsection{Finetuning RS-CapRet to Image Captioning datasets}

The results from the baselines MLCA-Net~\cite{cheng2022NWPUCaptions} and RSGPT~\cite{hu2023rsgpt} came from training or fine-tuning with the respective dataset. SkyEyeGPT~\cite{zhan2024skyeyegpt} has also evaluated their general model and finetuning it and observed that while the general model achieved high results, in some datasets finetuning would also help to improve the results. 
To experiment with the role of fine-tuning in the dataset on which the evaluation is done, we further fine-tuned RS-CapRet (with a learning rate of 0.0001), for each dataset separately. The results are collected in Table~\ref{table:cap-all-test} %
with the row \emph{finetuned} for each dataset. It can be seen that fine-tuning furthers the performance, and the results are accentuated in particular for the datasets of smaller size (UCM and Sydney-Captions). For NWPU, most metrics are improved, furthering even more the high performance of RS-CapRet in this dataset; as for RSICD the results suffered a slight degradation for all metrics; while for UCM and Sydney the results greatly improved. In UCM, RS-CapRet already had higher performance in different metrics compared to the baselines (METEOR, ROUGE\_L, CIDEr, SPICE) and with this fine-tuning the BLEU results (BLEU-2, BLEU-4) are now higher than the baselines as well. These experiments show that fine-tuning exclusively on which the dataset the evaluation is being done helps with the results, however, we argue the resulting model is less interesting since it is now specific for that dataset, and does not have the overall performance and generalization abilities that we are looking for.

\begin{figure*}[t!]
\centering
\includegraphics[width=\linewidth]{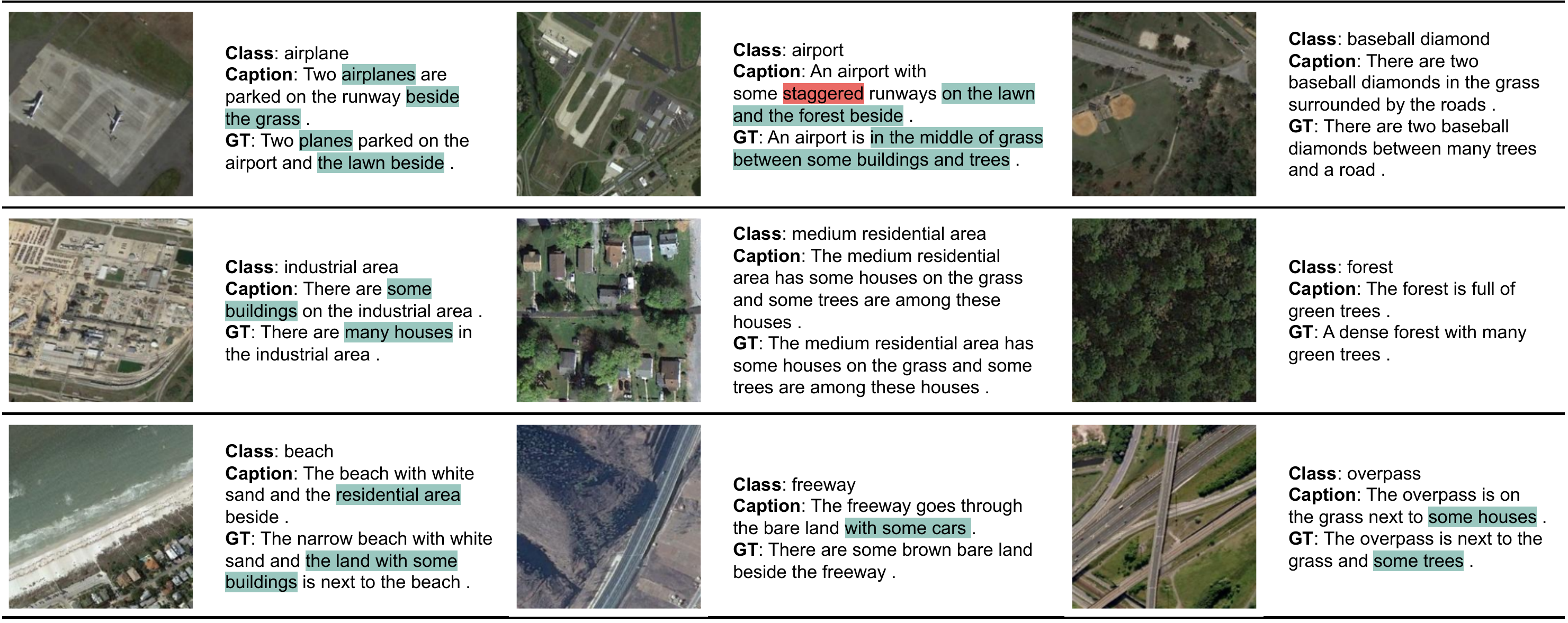}
\caption{Qualitative examples of generated captions given images of different classes of the test-set of NWPU-Captions dataset~\cite{cheng2022NWPUCaptions}.}
\label{fig:captions_examples}
\end{figure*}

\begin{figure*}[t!]
\centering
\includegraphics[width=0.7\linewidth]{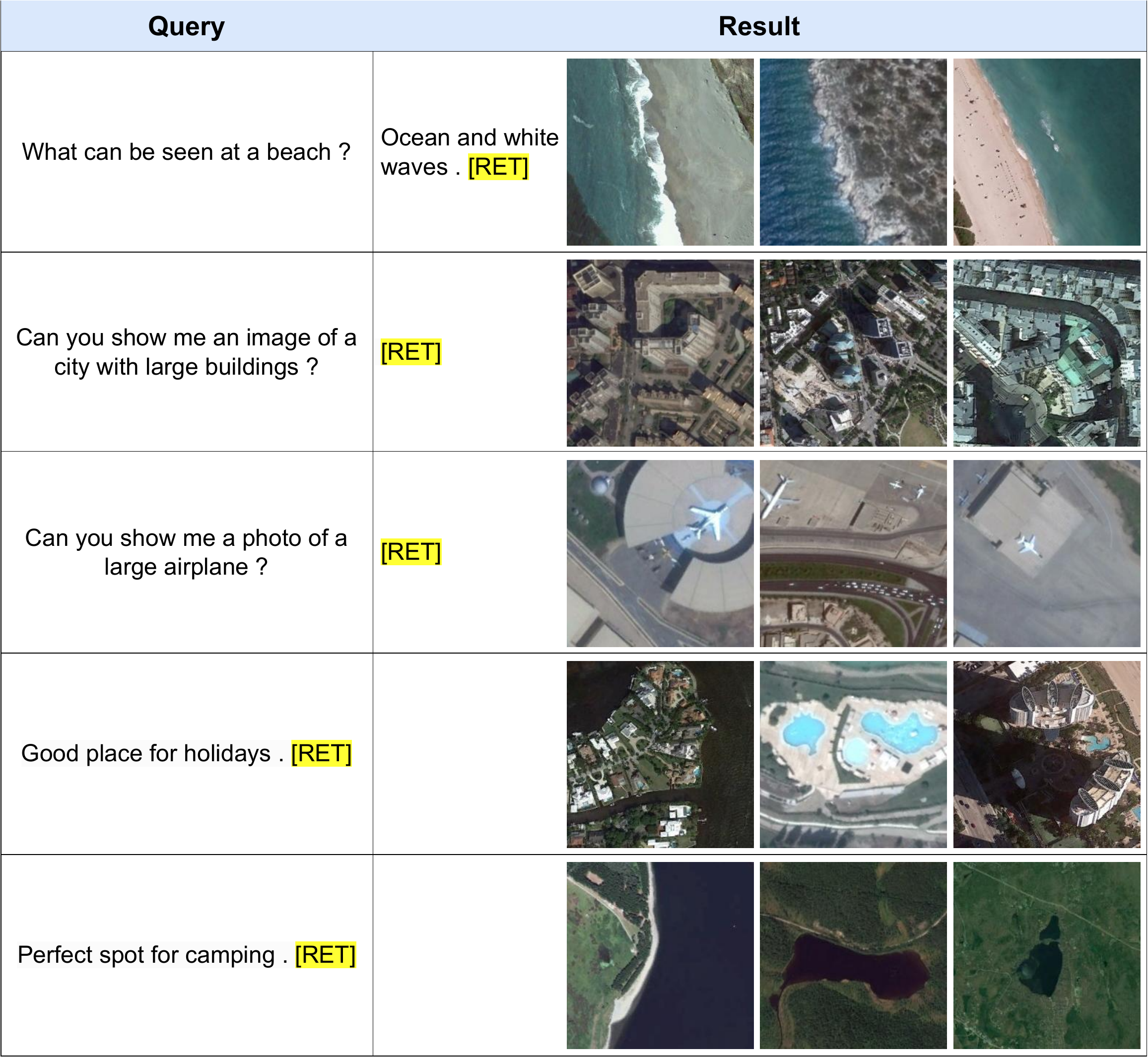}
\caption{Examples image retrieval by RS-CapRet given different requests by the user, considering object features and related topics.}
\label{fig:request}
\end{figure*}

\begin{figure*}[t!]
\centering
\includegraphics[width=\linewidth]{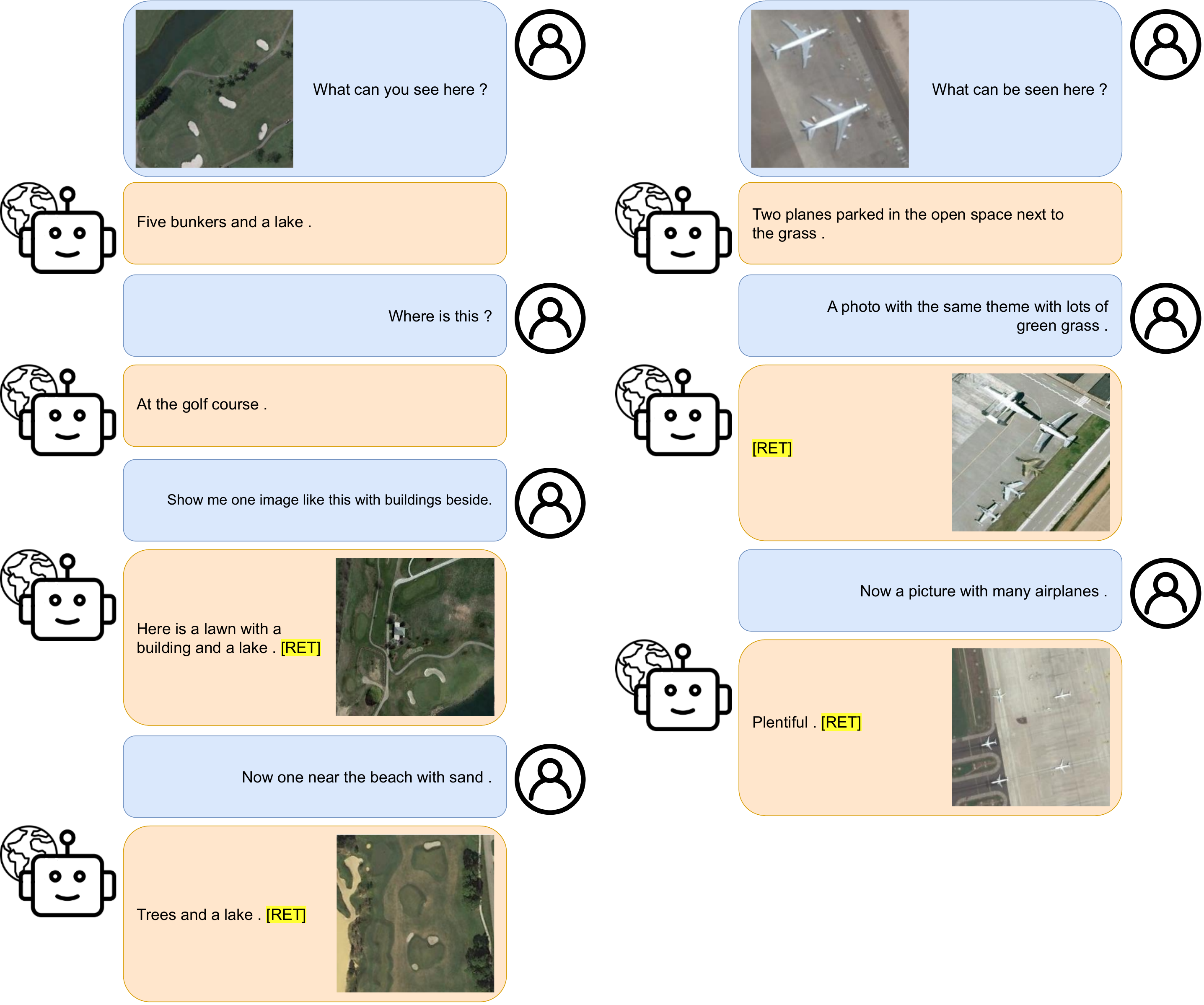}
\caption{Examples of dialogue with RS-CapRet, showing a) the ability to handle multi-modal inputs with interleaved sequences of images and text as well as b) reasoning abilities given world knowledge.}
\label{fig:dialogue}
\end{figure*}

\begin{figure*}[t!]
\centering
\includegraphics[width=0.66\linewidth]{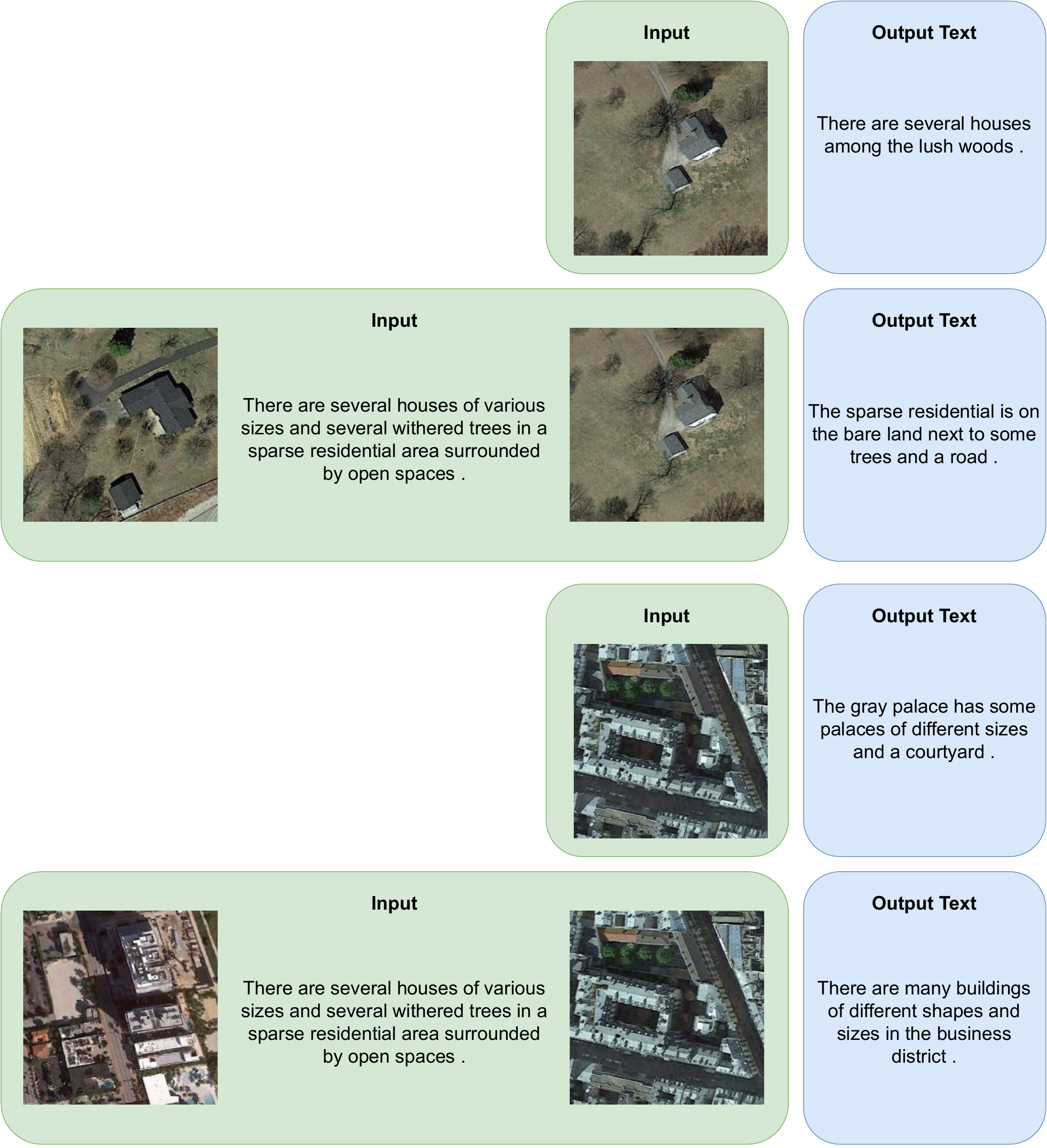}
\caption{In-context learning ability of RS-CapRet: Given one example of the correct class of the input image, RS-CapRet can generate an accurate description where it had before failed.}
\label{fig:icl}
\end{figure*}

\subsection{Retrieval Results}

We evaluated the performance of RS-CapRet for retrieval tasks in the RSICD and UCM datasets. These are both reported in Table~\ref{table:retrieval-ucm}. To assess the results achieved, we include baselines such as GaLR~\cite{yuan2022remote}, and KCR~\cite{mi2022knowledge}. We also include other works that have leveraged CLIP-based models such as Rahhal et al.~\cite{rahhal2022multilanguage} and RemoteCLIP~\cite{remoteclip}. CLIP results are also included with models already mentioned in subsection~\ref{subsection:clipchoice}: CLIP-ViT-B/32; CLIP-ViT-B/14; fine-tuned CLIP-ViT-B/32 in RSICD from~\cite{pal2021Fine}; and CLIP-Cap-4. We collect the results reported for the baselines, while for our CLIP models we evaluate them in our retrieval setup.

As previously discussed, the original CLIP weights of the model already obtain a reasonable performance, and a vision encoder ViT-L/14 obtains a better performance compared to a modal of size ViT-B/32. The finetuned version in RSICD of CLIP ViT-B/32 by Pal et al.~\cite{pal2021Fine} can further improve the results in both datasets. Our finetuned CLIP-Cap-4 data can further improve the results and be comparable with RemoteCLIP with a ViT-L backbone. RS-CapRet can achieve higher results when compared to GaLR, KCR, Rahhal et al, and also zero-shot CLIP of size Base and Large. However, it cannot surpass CLIP-RSICD~\cite{pal2021Fine} and larger variants fine-tuned to the remote sensing domain. 

For the UCM dataset, a similar pattern can be observed. RS-CapRet can achieve higher results compared to KCR, and zero-shot CLIPs of size Base and Large. Yet, the performance is lower compared to Rahhal et al.~\cite{rahhal2022multilanguage}, and RemoteCLIP~\cite{remoteclip}.

We note that CLIP is already a very strong baseline and that it can be further improved with fine-tuning in the domain, as observed in our experiments and related work~\cite{remoteclip,rahhal2022multilanguage}. It follows expectations that CLIP can be better for text-image retrieval than RS-CapRet, given that in CLIP there are two fully dedicated Transformer encoders to obtain features from the images and the text that match corresponding pairs, while in RS-CapRet only projection layers for the embedding of the special $\rettoken$ token and the visual embeddings are trained to address this task. Yet, RS-CapRet can obtain high results and be competitve. When further fine-tuning RS-CapRet on the datasets, performance is slightly improved for both datasets. %

\subsection{Usage of different types of Vision Encoder}

As mentioned in previous sections, the remote sensing community has already proposed vision encoders finetuned for the domain, mainly with models based on masked autoencoders or masked image modeling objectives~\cite{wang2022Advancing,sun2022Ringmo,cha2023billion}. We test the use of a publicly available ViT-B pre-trained on the MillionAID dataset, with a MAE objective which we refer to as RS-ViT-B~\cite{wang2022Advancing}. We compare this encoder with a CLIP-based model finetuned to the remote sensing domain as the ones we have been using but with the same ViT-B backbone, namely the CLIP-RSICD~\cite{pal2021Fine}. %
For this set of experiments, the decoder model is also LLamaV2-7b, and the model is still trained with the Cap-4 dataset. Image captioning and text-image retrieval results for these experiments are reported in Table~\ref{table:ablation-IC}. %
It can be seen that using a vision encoder based on CLIP in our method leads to better results in image captioning across the different datasets, as well as better results in text-image retrieval. CLIP-Cap-4 of larger size leads to better results on the overall model. These results follow findings from previous work which argue that the use of vision encoders that were pre-trained with a text supervision signal leads to better performance when the embeddings of these models are integrated with LLMs~\cite{koh2023Grounding,merullo2022linearly}.

\section{Qualitative Analysis}

Some examples of captions generated by RS-CapRet are illustrated in Fig.~\ref{fig:captions_examples}, where examples of different classes of the NWPU-Captions test set were chosen, namely \emph{airplane}, \emph{airport}, \emph{baseball field}, \emph{industrial area}, \emph{medium residential area}, \emph{forest}, \emph{beach}, \emph{freeway}, and \emph{overpass}. An example of a caption of the dataset that is most simlar to the generated caption is included. These examples show that the model can generate short descriptions of remote sensing images, following the format of the captions in the original dataset by describing a general overview of the image with a general description of the positional relationship between the most relevant objects.

An example of requests that can be made by the user can be seen in Fig.~\ref{fig:request}. For the different queries, the top-3 images with the most similar embeddings to the  $\rettoken$ token were retrieved. For the first three examples, a query was given to the model which generated the special $\rettoken$ or a short description before it. It can be seen that the model can obtain images considering the different types of scenes or objects requested, namely a beach, a city with tall buildings, and a large airplane. 

Due to the high performance of the LLM on which RS-CapRet is based on, RS-CapRet can leverage knowledge between concepts that it has learned about beyond our training. We experimented with asking the model to obtain images related to concepts, in particular images related to holidays and camping spots, illustrated in the two bottom examples of Fig.~\ref{fig:request}. For these examples, we directly appended the $\rettoken$ to the end of the text and obtained its representation from the model. In the example of the holidays, the images retrieved consist of pools and a hotel resort; regarding a spot for camping, the images obtained depict a beach and a lake in a forest.

The ability to integrate visual embeddings into the input space of the LLM allows further combination of multi-modal inputs, with sequences of multiple images along with text and questions. This was further enhanced during the training setup where two image and text sequences were concatenated, while the model was trained for the image captioning objective. We experimented with the ability of the model to handle short conversations given multimodal inputs. Some dialogue examples that can be done with RS-CapRet are illustrated in Fig.~\ref{fig:dialogue}. In the example on the left, the model receives as input an image together with a question and can generate a description that answers the question. A small conversation can be made regarding the generated caption. When asked to obtain a variation of the initial image to include buildings, without the explicit mention by the user that the scene described is a golf course ("\emph{Show me one...}"), the model can obtain an image that fits the desired criteria. A second variant can be requested, and an image that fits the desired criteria (a golf course near the beach with sand) can be obtained as well. A second example is also shown on the right in Fig.~\ref{fig:dialogue} regarding airport scenery, where the model can provide a description of the initial image, obtain a variant of the scenery, and another variant regarding the number of objects. Despite these promising results, we have observed that the model can also lack robustness due to not having been directly optimized for this type of task during training, still, these experiments show promising avenues for conversational agents for the remote sensing domain.

\section{Conclusions and Future Work}
In this work, we have proposed a Vision and Language model for the remote sensing domain named RS-CapRet that can address both tasks of image captioning as well as text-image retrieval. We obtain SOTA or competitive results with existing methods with a simple procedure of including visual embeddings of remote sensing images with currently available Large Language Models. RS-CapRet can also obtain images given user requests, given particular objects or related themes; RS-CapRet also shows it can handle conversations about remote sensing images, describing input images, and obtaining other given user queries. There are further avenues of work that can be addressed based on this work: extend the capabilities of the model for Vision and Language tasks by explicitly addressing visual question answering~\cite{lobry2020rsvqa} and visual grounding~\cite{chen2023shikra,kuckreja2023geochat,zhan2024skyeyegpt}, include conversational data from chat logs and instructions for more robust and high-quality interaction with the model~\cite{liu2023visual}, as well as data with more dense descriptions of remote sensing images~\cite{hu2023rsgpt}, with more details regarding attributes of objects and positional relationships between them. Finally, the architecture could be extended by including mechanisms that obtain image embeddings in a way that is more suited to the characteristics of remote sensing images~\cite{cheng2022NWPUCaptions}.

\section*{Acknowledgment}

This research was supported by the Portuguese Recovery and Resilience Plan through project C645008882-00000055 (i.e., the Center For Responsible AI), and also by Funda\c{c}\~ao para a Ci\^encia e Tecnologia (FCT), through the project with reference UIDB/50021/2020 (DOI:10.54499/UIDB/50021/2020).

\ifCLASSOPTIONcaptionsoff
  \newpage
\fi

\bibliographystyle{IEEEtran}
\bibliography{tidy}

\begin{thebibliography}{10}
\providecommand{\url}[1]{#1}
\csname url@samestyle\endcsname
\providecommand{\newblock}{\relax}
\providecommand{\bibinfo}[2]{#2}
\providecommand{\BIBentrySTDinterwordspacing}{\spaceskip=0pt\relax}
\providecommand{\BIBentryALTinterwordstretchfactor}{4}
\providecommand{\BIBentryALTinterwordspacing}{\spaceskip=\fontdimen2\font plus
\BIBentryALTinterwordstretchfactor\fontdimen3\font minus
  \fontdimen4\font\relax}
\providecommand{\BIBforeignlanguage}[2]{{%
\expandafter\ifx\csname l@#1\endcsname\relax
\typeout{** WARNING: IEEEtran.bst: No hyphenation pattern has been}%
\typeout{** loaded for the language `#1'. Using the pattern for}%
\typeout{** the default language instead.}%
\else
\language=\csname l@#1\endcsname
\fi
#2}}
\providecommand{\BIBdecl}{\relax}
\BIBdecl

\bibitem{Wen2023vision}
C.~Wen, Y.~Hu, X.~Li, Z.~Yuan, and X.~X. Zhu, ``Vision-language models in
  remote sensing: {{Current}} progress and future trends,''
  \emph{arXiv:2305.05726}, 2023.

\bibitem{mai2023opportunities}
G.~Mai, W.~Huang, J.~Sun, S.~Song, D.~Mishra, N.~Liu, S.~Gao, T.~Liu, G.~Cong,
  Y.~Hu \emph{et~al.}, ``On the opportunities and challenges of foundation
  models for geospatial artificial intelligence,'' \emph{arXiv:2304.06798},
  2023.

\bibitem{remoteclip}
F.~Liu, D.~Chen, Z.~Guan, X.~Zhou, J.~Zhu, and J.~Zhou, ``Remoteclip: {A}
  vision language foundation model for remote sensing,''
  \emph{arXiv:2306.11029}, 2023.

\bibitem{rahhal2022multilanguage}
M.~M.~A. Rahhal, Y.~Bazi, N.~A. Alsharif, L.~Bashmal, N.~Alajlan, and
  F.~Melgani, ``Multilanguage transformer for improved text to remote sensing
  image retrieval,'' \emph{IEEE Journal of Selected Topics in Applied Earth
  Observations and Remote Sensing}, vol.~15, pp. 9115--9126, 2022.

\bibitem{mi2022knowledge}
L.~Mi, S.~Li, C.~Chappuis, and D.~Tuia, ``Knowledge-aware cross-modal
  text-image retrieval for remote sensing images,'' in \emph{Proceedings of the
  Second Workshop on Complex Data Challenges in Earth Observation (CDCEO
  2022)}, 2022.

\bibitem{yuan2022remote}
Z.~Yuan, W.~Zhang, C.~Tian, X.~Rong, Z.~Zhang, H.~Wang, K.~Fu, and X.~Sun,
  ``Remote sensing cross-modal text-image retrieval based on global and local
  information,'' \emph{IEEE Transactions on Geoscience and Remote Sensing},
  vol.~60, pp. 1--16, 2022.

\bibitem{cheng2022NWPUCaptions}
Q.~Cheng, H.~Huang, Y.~Xu, Y.~Zhou, H.~Li, and Z.~Wang, ``Nwpu-captions dataset
  and mlca-net for remote sensing image captioning,'' \emph{IEEE Transactions
  on Geoscience and Remote Sensing}, vol.~60, pp. 1--19, 2022.

\bibitem{wei2023vlca}
T.~Wei, W.~Yuan, J.~Luo, W.~Zhang, and L.~Lu, ``{{VLCA}}: Vision-language
  aligning model with cross-modal attention for bilingual remote sensing image
  captioning,'' \emph{Journal of Systems Engineering and Electronics}, vol.~34,
  no.~1, pp. 9--18, 2023.

\bibitem{ramos2022using}
R.~Ramos and B.~Martins, ``Using {{Neural Encoder-Decoder Models With
  Continuous Outputs}} for {{Remote Sensing Image Captioning}},'' \emph{IEEE
  Access}, vol.~10, 2022.

\bibitem{shi2017Can}
Z.~Shi and Z.~Zou, ``Can a {{Machine Generate Humanlike Language Descriptions}}
  for a {{Remote Sensing Image}}?'' \emph{IEEE Transactions on Geoscience and
  Remote Sensing}, vol.~55, 2017.

\bibitem{lobry2020rsvqa}
S.~Lobry, D.~Marcos, J.~Murray, and D.~Tuia, ``{{RSVQA}}: {{Visual Question
  Answering}} for {{Remote Sensing Data}},'' \emph{IEEE Transactions on
  Geoscience and Remote Sensing}, vol.~58, 2020.

\bibitem{silva2022remote}
J.~D. Silva, J.~Magalh{\~a}es, D.~Tuia, and B.~Martins, ``Remote sensing visual
  question answering with a self-attention multi-modal encoder,'' in
  \emph{Proceedings of the 5th {{ACM SIGSPATIAL International Workshop}} on
  {{AI}} for {{Geographic Knowledge Discovery}}}.\hskip 1em plus 0.5em minus
  0.4em\relax {Association for Computing Machinery}, 2022, pp. 40--49.

\bibitem{bazi2022bi}
Y.~Bazi, M.~M. Al~Rahhal, M.~L. Mekhalfi, M.~A. Al~Zuair, and F.~Melgani,
  ``Bi-modal transformer-based approach for visual question answering in remote
  sensing imagery,'' \emph{IEEE Transactions on Geoscience and Remote Sensing},
  vol.~60, pp. 1--11, 2022.

\bibitem{zhang2023spatial}
Z.~Zhang, L.~Jiao, L.~Li, X.~Liu, P.~Chen, F.~Liu, Y.~Li, and Z.~Guo, ``A
  spatial hierarchical reasoning network for remote sensing visual question
  answering,'' \emph{IEEE Transactions on Geoscience and Remote Sensing},
  vol.~61, pp. 1--15, 2023.

\bibitem{tuia2021collective}
D.~Tuia, R.~Roscher, J.~D. Wegner, N.~Jacobs, X.~Zhu, and G.~{Camps-Valls},
  ``Toward a collective agenda on {{AI}} for earth science data analysis,''
  \emph{IEEE Geoscience and Remote Sensing Magazine}, vol.~9, no.~2, 2021.

\bibitem{martins2022towards}
B.~Martins and J.~D. Silva, ``Towards natural language interfaces for
  interacting with remote sensing data,'' \emph{UC Santa Barbara: Center for
  Spatial Studies}, 2022.

\bibitem{hu2023rsgpt}
Y.~Hu, J.~Yuan, C.~Wen, X.~Lu, and X.~Li, ``Rsgpt: A remote sensing vision
  language model and benchmark,'' \emph{arXiv:2307.15266}, 2023.

\bibitem{zhan2024skyeyegpt}
Y.~Zhan, Z.~Xiong, and Y.~Yuan, ``Skyeyegpt: Unifying remote sensing
  vision-language tasks via instruction tuning with large language model,''
  \emph{arXiv:2401.09712}, 2024.

\bibitem{kuckreja2023geochat}
K.~Kuckreja, M.~S. Danish, M.~Naseer, A.~Das, S.~Khan, and F.~S. Khan,
  ``Geochat: Grounded large vision-language model for remote sensing,''
  \emph{arXiv:2311.15826}, 2023.

\bibitem{wang2021SimVLM}
Z.~Wang, J.~Yu, A.~W. Yu, Z.~Dai, Y.~Tsvetkov, and Y.~Cao, ``{{SimVLM}}:
  {{Simple Visual Language Model Pretraining}} with {{Weak Supervision}},''
  \emph{arXiv:2108.10904}, 2021.

\bibitem{li2022blip}
J.~Li, D.~Li, C.~Xiong, and S.~Hoi, ``Blip: Bootstrapping language-image
  pre-training for unified vision-language understanding and generation,'' in
  \emph{International Conference on Machine Learning}.\hskip 1em plus 0.5em
  minus 0.4em\relax PMLR, 2022, pp. 12\,888--12\,900.

\bibitem{alayrac2022flamingo}
J.-B. Alayrac, J.~Donahue, P.~Luc, A.~Miech, I.~Barr, Y.~Hasson, K.~Lenc,
  A.~Mensch, K.~Millican, M.~Reynolds \emph{et~al.}, ``Flamingo: A visual
  language model for few-shot learning,'' \emph{Advances in Neural Information
  Processing Systems}, vol.~35, pp. 23\,716--23\,736, 2022.

\bibitem{hochreiter1997Long}
S.~Hochreiter and J.~Schmidhuber, ``Long {{Short-term Memory}},'' \emph{Neural
  computation}, vol.~9, pp. 1735--1780, 1997.

\bibitem{devlin2019BERT}
J.~Devlin, M.-W. Chang, K.~Lee, and K.~Toutanova, ``{{BERT}}: {{Pre-training}}
  of {{Deep Bidirectional Transformers}} for {{Language Understanding}},''
  \emph{arXiv:1810.04805}, 2019.

\bibitem{chappuis2022prompt}
C.~Chappuis, V.~Zermatten, S.~Lobry, B.~Le~Saux, and D.~Tuia, ``Prompt-rsvqa:
  Prompting visual context to a language model for remote sensing visual
  question answering,'' in \emph{Proceedings of the IEEE/CVF Conference on
  Computer Vision and Pattern Recognition}, 2022, pp. 1372--1381.

\bibitem{lu2017exploring}
X.~Lu, B.~Wang, X.~Zheng, and X.~Li, ``Exploring models and data for remote
  sensing image caption generation,'' \emph{IEEE Transactions on Geoscience and
  Remote Sensing}, vol.~56, 2018.

\bibitem{zia2022Transforming}
U.~Zia, M.~M. Riaz, and A.~Ghafoor, ``Transforming remote sensing images to
  textual descriptions,'' \emph{International Journal of Applied Earth
  Observation and Geoinformation}, vol. 108, p. 102741, 2022.

\bibitem{huang2021denoising}
W.~Huang, Q.~Wang, and X.~Li, ``Denoising-based multiscale feature fusion for
  remote sensing image captioning,'' \emph{IEEE Geoscience and Remote Sensing
  Letters}, vol.~18, no.~3, pp. 436--440, 2021.

\bibitem{li2020multi}
Y.~Li, S.~Fang, L.~Jiao, R.~Liu, and R.~Shang, ``A multi-level attention model
  for remote sensing image captions,'' \emph{Remote Sensing}, vol.~12, no.~6,
  2020.

\bibitem{yuan2020exploring}
Z.~Yuan, X.~Li, and Q.~Wang, ``Exploring multi-level attention and semantic
  relationship for remote sensing image captioning,'' \emph{IEEE Access},
  vol.~8, pp. 2608--2620, 2020.

\bibitem{zhao2021Highresolution}
R.~Zhao, Z.~Shi, and Z.~Zou, ``High-resolution remote sensing image captioning
  based on structured attention,'' \emph{IEEE Transactions on Geoscience and
  Remote Sensing}, vol.~60, pp. 1--14, 2021.

\bibitem{petroni2019language}
F.~Petroni, T.~Rockt{\"a}schel, P.~Lewis, A.~Bakhtin, Y.~Wu, A.~H. Miller, and
  S.~Riedel, ``Language models as knowledge bases?'' \emph{arXiv:1909.01066},
  2019.

\bibitem{brown2020language}
T.~Brown, B.~Mann, N.~Ryder, M.~Subbiah, J.~D. Kaplan, P.~Dhariwal,
  A.~Neelakantan, P.~Shyam, G.~Sastry, A.~Askell \emph{et~al.}, ``Language
  models are few-shot learners,'' \emph{Advances in neural information
  processing systems}, vol.~33, pp. 1877--1901, 2020.

\bibitem{wei2022emergent}
J.~Wei, Y.~Tay, R.~Bommasani, C.~Raffel, B.~Zoph, S.~Borgeaud, D.~Yogatama,
  M.~Bosma, D.~Zhou, D.~Metzler \emph{et~al.}, ``Emergent abilities of large
  language models,'' \emph{arXiv:2206.07682}, 2022.

\bibitem{wei2022chain}
J.~Wei, X.~Wang, D.~Schuurmans, M.~Bosma, F.~Xia, E.~Chi, Q.~V. Le, D.~Zhou
  \emph{et~al.}, ``Chain-of-thought prompting elicits reasoning in large
  language models,'' \emph{Advances in Neural Information Processing Systems},
  vol.~35, pp. 24\,824--24\,837, 2022.

\bibitem{kojima2022large}
T.~Kojima, S.~S. Gu, M.~Reid, Y.~Matsuo, and Y.~Iwasawa, ``Large language
  models are zero-shot reasoners,'' \emph{Advances in neural information
  processing systems}, vol.~35, pp. 22\,199--22\,213, 2022.

\bibitem{huang2022towards}
J.~Huang and K.~C.-C. Chang, ``Towards reasoning in large language models: A
  survey,'' \emph{arXiv:2212.10403}, 2022.

\bibitem{guo2022images}
J.~Guo, J.~Li, D.~Li, A.~M.~H. Tiong, B.~Li, D.~Tao, and S.~C. Hoi, ``From
  images to textual prompts: {{Zero-shot VQA}} with frozen large language
  models,'' \emph{arXiv:2212.10846}, 2022.

\bibitem{yang2022empirical}
Z.~Yang, Z.~Gan, J.~Wang, X.~Hu, Y.~Lu, Z.~Liu, and L.~Wang, ``An empirical
  study of gpt-3 for few-shot knowledge-based vqa,'' in \emph{Proceedings of
  the {{AAAI}} Conference on Artificial Intelligence}, vol.~36, 2022, pp.
  3081--3089.

\bibitem{tsimpoukelli2021Multimodal}
M.~Tsimpoukelli, J.~L. Menick, S.~Cabi, S.~Eslami, O.~Vinyals, and F.~Hill,
  ``Multimodal few-shot learning with frozen language models,'' \emph{Advances
  in Neural Information Processing Systems}, vol.~34, pp. 200--212, 2021.

\bibitem{hu2021lora}
E.~J. Hu, Y.~Shen, P.~Wallis, Z.~Allen-Zhu, Y.~Li, S.~Wang, L.~Wang, and
  W.~Chen, ``Lora: Low-rank adaptation of large language models,''
  \emph{arXiv:2106.09685}, 2021.

\bibitem{lester2021power}
B.~Lester, R.~Al{-}Rfou, and N.~Constant, ``The power of scale for
  parameter-efficient prompt tuning,'' in \emph{Proceedings of the 2021
  Conference on Empirical Methods in Natural Language Processing, {EMNLP} 2021,
  Virtual Event / Punta Cana, Dominican Republic, 7-11 November, 2021}.\hskip
  1em plus 0.5em minus 0.4em\relax Association for Computational Linguistics,
  2021, pp. 3045--3059.

\bibitem{koh2023Grounding}
J.~Y. Koh, R.~Salakhutdinov, and D.~Fried, ``Grounding language models to
  images for multimodal inputs and outputs,'' in \emph{International Conference
  on Machine Learning}.\hskip 1em plus 0.5em minus 0.4em\relax PMLR, 2023, pp.
  17\,283--17\,300.

\bibitem{eichenberg2022MAGMA}
\BIBentryALTinterwordspacing
C.~Eichenberg, S.~Black, S.~Weinbach, L.~Parcalabescu, and A.~Frank, ``{MAGMA}
  {--} multimodal augmentation of generative models through adapter-based
  finetuning,'' in \emph{Findings of the Association for Computational
  Linguistics: EMNLP 2022}, Y.~Goldberg, Z.~Kozareva, and Y.~Zhang, Eds.\hskip
  1em plus 0.5em minus 0.4em\relax Abu Dhabi, United Arab Emirates: Association
  for Computational Linguistics, Dec. 2022, pp. 2416--2428. [Online].
  Available: \url{https://aclanthology.org/2022.findings-emnlp.179}
\BIBentrySTDinterwordspacing

\bibitem{ramos2022smallcap}
R.~Ramos, B.~Martins, D.~Elliott, and Y.~Kementchedjhieva, ``Smallcap:
  lightweight image captioning prompted with retrieval augmentation,'' pp.
  2840--2849, 2023.

\bibitem{li2023blip}
J.~Li, D.~Li, S.~Savarese, and S.~Hoi, ``Blip-2: bootstrapping language-image
  pre-training with frozen image encoders and large language models,'' in
  \emph{Proceedings of the 40th International Conference on Machine Learning},
  ser. ICML'23.\hskip 1em plus 0.5em minus 0.4em\relax JMLR.org, 2023.

\bibitem{qu2016Deep}
B.~Qu, X.~Li, D.~Tao, and X.~Lu, ``Deep semantic understanding of high
  resolution remote sensing image,'' in \emph{Proceedings of the
  {{International Conference}} on {{Computer}}, {{Information}} and
  {{Telecommunication Systems}}}, 2016.

\bibitem{sumbul2021sdrsic}
G.~Sumbul, S.~Nayak, and B.~Demir, ``Sd-rsic: Summarization-driven deep remote
  sensing image captioning,'' \emph{IEEE Transactions on Geoscience and Remote
  Sensing}, vol.~59, no.~8, pp. 6922--6934, 2021.

\bibitem{li2021truncation}
X.~Li, X.~Zhang, W.~Huang, and Q.~Wang, ``Truncation cross entropy loss for
  remote sensing image captioning,'' \emph{IEEE Transactions on Geoscience and
  Remote Sensing}, vol.~59, no.~6, pp. 5246--5257, 2021.

\bibitem{Simonyan2014vgg}
K.~Simonyan and A.~Zisserman, ``Very deep convolutional networks for
  large-scale image recognition,'' \emph{arXiv:1409.1556}, 2014.

\bibitem{vaswani2017Attention}
A.~Vaswani, N.~Shazeer, N.~Parmar, J.~Uszkoreit, L.~Jones, A.~N. Gomez,
  {\L}.~Kaiser, and I.~Polosukhin, ``Attention is all you need,''
  \emph{Advances in neural information processing systems}, vol.~30, 2017.

\bibitem{radford2019Language}
A.~Radford, J.~Wu, R.~Child, D.~Luan, D.~Amodei, I.~Sutskever \emph{et~al.},
  ``Language models are unsupervised multitask learners,'' \emph{Open AI Blog},
  2019.

\bibitem{reimers2019SentenceBERT}
N.~Reimers and I.~Gurevych, ``Sentence-{BERT}: Sentence embeddings using
  {S}iamese {BERT}-networks,'' in \emph{Proceedings of the 2019 Conference on
  Empirical Methods in Natural Language Processing and the 9th International
  Joint Conference on Natural Language Processing (EMNLP-IJCNLP)}, K.~Inui,
  J.~Jiang, V.~Ng, and X.~Wan, Eds.\hskip 1em plus 0.5em minus 0.4em\relax Hong
  Kong, China: Association for Computational Linguistics, Nov. 2019, pp.
  3982--3992.

\bibitem{radford2021Learninga}
A.~Radford, J.~W. Kim, C.~Hallacy, A.~Ramesh, G.~Goh, S.~Agarwal, G.~Sastry,
  A.~Askell, P.~Mishkin, J.~Clark, G.~Krueger, and I.~Sutskever, ``Learning
  {{Transferable Visual Models From Natural Language Supervision}},'' in
  \emph{Proceedings of the 38th {{International Conference}} on {{Machine
  Learning}}, {{ICML}} 2021, 18-24 {{July}} 2021, {{Virtual Event}}}, vol.
  139.\hskip 1em plus 0.5em minus 0.4em\relax {PMLR}, 2021, pp. 8748--8763.

\bibitem{pal2021Fine}
S.~Pal, A.~Arutiunian, G.~Venkatesh, R.~Ghosh, D.~Vidhani, and M.~Bhaskar,
  ``Fine tuning {{CLIP}} with {{Remote Sensing}} ({{Satellite}}) images and
  captions,'' \emph{HuggingFace Blog}, 2021.

\bibitem{dosovitskiy2020vit}
A.~Dosovitskiy, L.~Beyer, A.~Kolesnikov, D.~Weissenborn, X.~Zhai,
  T.~Unterthiner, M.~Dehghani, M.~Minderer, G.~Heigold, S.~Gelly \emph{et~al.},
  ``An image is worth 16x16 words: Transformers for image recognition at
  scale,'' \emph{arXiv:2010.11929}, 2020.

\bibitem{zermatten2023tacoss}
V.~Zermatten, J.~C. Navarro, L.~Hughes, T.~Kellenberger, and D.~Tuia, ``Text as
  a richer source of supervision in semantic segmentation tasks,'' in
  \emph{IGARSS 2023 - 2023 IEEE International Geoscience and Remote Sensing
  Symposium}, 2023, pp. 2219--2222.

\bibitem{jakubik2023toward}
J.~Jakubik, M.~Muszynski, M.~V{\"o}ssing, N.~K{\"u}hl, and T.~Brunschwiler,
  ``Toward foundation models for earth monitoring: {{Generalizable}} deep
  learning models for natural hazard segmentation,'' \emph{arXiv:2301.09318},
  2023.

\bibitem{wang2022Advancing}
D.~Wang, Q.~Zhang, Y.~Xu, J.~Zhang, B.~Du, D.~Tao, and L.~Zhang, ``Advancing
  plain vision transformer towards remote sensing foundation model,''
  \emph{IEEE Transactions on Geoscience and Remote Sensing}, 2022.

\bibitem{he2022Masked}
K.~He, X.~Chen, S.~Xie, Y.~Li, P.~Doll{\'a}r, and R.~B. Girshick, ``Masked
  {{Autoencoders Are Scalable Vision Learners}},'' in \emph{{{IEEE}}/{{CVF
  Conference}} on {{Computer Vision}} and {{Pattern Recognition}}, {{CVPR}}
  2022, {{New Orleans}}, {{LA}}, {{USA}}, {{June}} 18-24, 2022}.\hskip 1em plus
  0.5em minus 0.4em\relax {IEEE}, 2022, pp. 15\,979--15\,988.

\bibitem{Long2021DiRS}
Y.~Long, G.-S. Xia, S.~Li, W.~Yang, M.~Y. Yang, X.~X. Zhu, L.~Zhang, and D.~Li,
  ``On creating benchmark dataset for aerial image interpretation: Reviews,
  guidances and million-aid,'' \emph{IEEE Journal of Selected Topics in Applied
  Earth Observations and Remote Sensing}, vol.~14, pp. 4205--4230, 2021.

\bibitem{sun2022Ringmo}
X.~Sun, P.~Wang, W.~Lu, Z.~Zhu, X.~Lu, Q.~He, J.~Li, X.~Rong, Z.~Yang, H.~Chang
  \emph{et~al.}, ``Ringmo: {{A}} remote sensing foundation model with masked
  image modeling,'' \emph{IEEE Transactions on Geoscience and Remote Sensing},
  2022.

\bibitem{cha2023billion}
K.~Cha, J.~Seo, and T.~Lee, ``A billion-scale foundation model for remote
  sensing images,'' \emph{arXiv:2304.05215}, 2023.

\bibitem{mendieta2023gfm}
M.~Mendieta, B.~Han, X.~Shi, Y.~Zhu, C.~Chen, and M.~Li, ``{{GFM}}:
  {{Building}} geospatial foundation models via continual pretraining,''
  \emph{arXiv:2302.04476}, 2023.

\bibitem{lacoste2023geobench}
A.~Lacoste, N.~Lehmann, P.~Rodriguez, E.~D. Sherwin, H.~Kerner, B.~L{\"u}tjens,
  J.~A. Irvin, D.~Dao, H.~Alemohammad, A.~Drouin, M.~Gunturkun, G.~Huang,
  D.~Vazquez, D.~Newman, Y.~Bengio, S.~Ermon, and X.~X. Zhu, ``{GEO}-bench:
  Toward foundation models for earth monitoring,'' in \emph{Thirty-seventh
  Conference on Neural Information Processing Systems Datasets and Benchmarks
  Track}, 2023.

\bibitem{dai2023instructblip}
W.~Dai, J.~Li, D.~Li, A.~M.~H. Tiong, J.~Zhao, W.~Wang, B.~Li, P.~Fung, and
  S.~Hoi, ``Instructblip: Towards general-purpose vision-language models with
  instruction tuning,'' \emph{arXiv:2305.06500}, 2023.

\bibitem{liu2023improved}
H.~Liu, C.~Li, Y.~Li, and Y.~J. Lee, ``Improved baselines with visual
  instruction tuning,'' \emph{arXiv:2310.03744}, 2023.

\bibitem{chen2023minigpt}
J.~Chen, D.~Zhu, X.~Shen, X.~Li, Z.~Liu, P.~Zhang, R.~Krishnamoorthi,
  V.~Chandra, Y.~Xiong, and M.~Elhoseiny, ``Minigpt-v2: large language model as
  a unified interface for vision-language multi-task learning,''
  \emph{arXiv:2310.09478}, 2023.

\bibitem{touvron2023llama2}
H.~Touvron, L.~Martin, K.~Stone, P.~Albert, A.~Almahairi, Y.~Babaei,
  N.~Bashlykov, S.~Batra, P.~Bhargava, S.~Bhosale \emph{et~al.}, ``Llama 2:
  Open foundation and fine-tuned chat models,'' \emph{arXiv:2307.09288}, 2023.

\bibitem{liu2023visual}
H.~Liu, C.~Li, Q.~Wu, and Y.~J. Lee, ``Visual instruction tuning,''
  \emph{arXiv:2304.08485}, 2023.

\bibitem{roberts2020How}
A.~Roberts, C.~Raffel, and N.~Shazeer, ``How much knowledge can you pack into
  the parameters of a language model?'' \emph{arXiv:2002.08910}, 2020.

\bibitem{ganguli2022Predictability}
D.~Ganguli, D.~Hernandez, L.~Lovitt, A.~Askell, Y.~Bai, A.~Chen, T.~Conerly,
  N.~Dassarma, D.~Drain, N.~Elhage \emph{et~al.}, ``Predictability and surprise
  in large generative models,'' in \emph{2022 {{ACM Conference}} on
  {{Fairness}}, {{Accountability}}, and {{Transparency}}}, 2022, pp.
  1747--1764.

\bibitem{jia2021scalinga}
C.~Jia, Y.~Yang, Y.~Xia, Y.-T. Chen, Z.~Parekh, H.~Pham, Q.~Le, Y.-H. Sung,
  Z.~Li, and T.~Duerig, ``Scaling up visual and vision-language representation
  learning with noisy text supervision,'' in \emph{International Conference on
  Machine Learning}.\hskip 1em plus 0.5em minus 0.4em\relax PMLR, 2021, pp.
  4904--4916.

\bibitem{fang2023eva}
Y.~Fang, W.~Wang, B.~Xie, Q.~Sun, L.~Wu, X.~Wang, T.~Huang, X.~Wang, and
  Y.~Cao, ``Eva: Exploring the limits of masked visual representation learning
  at scale,'' in \emph{Proceedings of the IEEE/CVF Conference on Computer
  Vision and Pattern Recognition}, 2023, pp. 19\,358--19\,369.

\bibitem{vicuna2023}
\BIBentryALTinterwordspacing
W.-L. Chiang, Z.~Li, Z.~Lin, Y.~Sheng, Z.~Wu, H.~Zhang, L.~Zheng, S.~Zhuang,
  Y.~Zhuang, J.~E. Gonzalez, I.~Stoica, and E.~P. Xing, ``Vicuna: An
  open-source chatbot impressing gpt-4 with 90\%* chatgpt quality,'' March
  2023. [Online]. Available: \url{https://lmsys.org/blog/2023-03-30-vicuna/}
\BIBentrySTDinterwordspacing

\bibitem{touvron2023LLaMA}
H.~Touvron, T.~Lavril, G.~Izacard, X.~Martinet, M.-A. Lachaux, T.~Lacroix,
  B.~Rozi{\`e}re, N.~Goyal, E.~Hambro, F.~Azhar, A.~Rodriguez, A.~Joulin,
  E.~Grave, and G.~Lample, ``{{LLaMA}}: {{Open}} and {{Efficient Foundation
  Language Models}},'' \emph{arXiv:2302.13971}, 2023.

\bibitem{hoffmann2022training}
J.~Hoffmann, S.~Borgeaud, A.~Mensch, E.~Buchatskaya, T.~Cai, E.~Rutherford,
  D.~d.~L. Casas, L.~A. Hendricks, J.~Welbl, A.~Clark \emph{et~al.}, ``Training
  compute-optimal large language models,'' \emph{arXiv:2203.15556}, 2022.

\bibitem{paszke2019pytorch}
A.~Paszke, S.~Gross, F.~Massa, A.~Lerer, J.~Bradbury, G.~Chanan, T.~Killeen,
  Z.~Lin, N.~Gimelshein, L.~Antiga \emph{et~al.}, ``Pytorch: An imperative
  style, high-performance deep learning library,'' \emph{Advances in neural
  information processing systems}, vol.~32, 2019.

\bibitem{chen2020Simple}
T.~Chen, S.~Kornblith, M.~Norouzi, and G.~Hinton, ``A {{Simple Framework}} for
  {{Contrastive Learning}} of {{Visual Representations}},''
  \emph{arXiv:2002.05709}, 2020.

\bibitem{merullo2022linearly}
J.~Merullo, L.~Castricato, C.~Eickhoff, and E.~Pavlick, ``Linearly mapping from
  image to text space,'' \emph{arXiv:2209.15162}, 2022.

\bibitem{chen2023shikra}
K.~Chen, Z.~Zhang, W.~Zeng, R.~Zhang, F.~Zhu, and R.~Zhao, ``Shikra: Unleashing
  multimodal llm's referential dialogue magic,'' \emph{arXiv:2306.15195}, 2023.

\end{thebibliography}

\end{document}